\def\ARXIV{}
\ifdefined\ARXIV
  \documentclass[doc,floatsintext,natbib]{apa7}
\else
  \documentclass[man,floatsintext,natbib]{apa7}
\fi

\usepackage{amsmath}
\usepackage{array}

\newcommand{\tighttable}{\linespread{1}\small\selectfont}

\newcommand{\valNComplete}{136}               
\newcommand{\valNAware}{23}                   
\newcommand{\valNAnalyzed}{113}               
\newcommand{\valNTrials}{9{,}040}             
\newcommand{\valNNewRun}{115}                 
\newcommand{\valNReused}{21}                  
\newcommand{\valNSampleTrials}{10{,}880}      
\newcommand{\valNItems}{8}                    
\newcommand{\valNReps}{5}                     
\newcommand{\valNCond}{2}                     
\newcommand{\valNTrialsPerModel}{80}          
\newcommand{\valNTrialsPerCell}{40}           
\newcommand{\valNBoot}{5{,}000}               
\newcommand{\valNOrgs}{39}                    
\newcommand{\valOrgList}{Abacus.AI, AI Singapore, AI9Stars, AIDC-AI, Alibaba (Qwen), Arcee AI, Baichuan, Cohere, deepreinforce-ai, DeepSeek, Dnotitia, Gnani.ai, Google, iApp Technology, IBM, InternLM, KAIST, Kwaipilot, LG AI Research, Mistral AI, Moonshot AI, Naver, nex-agi, NVIDIA, OpenBMB, poolside, Ruadapt, SAAI, T-Technologies, Tencent, Thinking Machines Lab, TokenRhythm, Trendyol, XHToken, Xiaohongshu, Xiaomi, YannQi, Z.AI, and Zhihu}
\newcommand{\valNexclManip}{9}                
\newcommand{\valNtrueOnlyAn}{108}             
\newcommand{\valNsettingsAn}{5}               
\newcommand{\valNNewRunAn}{92}                
\newcommand{\valNvllmNew}{75}                 
\newcommand{\valNllamaNew}{17}                
\newcommand{\valNkeyEnableThinking}{95}       
\newcommand{\valNkeyThinking}{8}              
\newcommand{\valNkeyOther}{10}                
\newcommand{\valManipReasonMin}{95}           
\newcommand{\valManipIntuMax}{13}             
\newcommand{\valManipRatioMin}{5.6}           
\newcommand{\valNattemptTwo}{9}               
\newcommand{\valNattemptThree}{2}             
\newcommand{\valNTruncSample}{17}             
\newcommand{\valNTruncModels}{1}              
\newcommand{\valTokIntMedian}{9}              
\newcommand{\valTokReaMedian}{389}            
\newcommand{\valParamMin}{0.5}                
\newcommand{\valParamMax}{1{,}020}            
\newcommand{\valParamMedian}{31}              
\newcommand{\valHumanN}{4{,}407}              
\newcommand{\valScanModels}{136}              
\newcommand{\valScanTrials}{10{,}880}         
\newcommand{\valScanCharsM}{16.3}             
\newcommand{\valScanAware}{23}                
\newcommand{\valScanAwarePct}{16.9}           
\newcommand{\valSigA}{5}                      
\newcommand{\valSigB}{8}                      
\newcommand{\valSigC}{0}                      
\newcommand{\valSigD}{18}                     
\newcommand{\valSigE}{2}                      
\newcommand{\valStructTrick}{36}              
\newcommand{\valStructTemplate}{34}           
\newcommand{\valStructLure}{25}               
\newcommand{\valStructOnly}{47}               
\newcommand{\valAccIntMean}{.832}             
\newcommand{\valAccIntLo}{.787}               
\newcommand{\valAccIntHi}{.872}               
\newcommand{\valAccReaMean}{.987}             
\newcommand{\valAccReaLo}{.973}               
\newcommand{\valAccReaHi}{.996}               
\newcommand{\valAccDeltaMean}{+.155}          
\newcommand{\valAccDeltaMedian}{+.100}        
\newcommand{\valAccDeltaLo}{+.119}            
\newcommand{\valAccDeltaHi}{+.193}            
\newcommand{\valAccNPos}{67}                  
\newcommand{\valAccSignP}{< .001}             
\newcommand{\valAccWilcoxonP}{< .001}         
\newcommand{\valLureIntMean}{.090}            
\newcommand{\valLureIntLo}{.068}              
\newcommand{\valLureIntHi}{.115}              
\newcommand{\valLureReaMean}{.003}            
\newcommand{\valLureReaLo}{.001}              
\newcommand{\valLureReaHi}{.006}              
\newcommand{\valLureDeltaMean}{-.087}         
\newcommand{\valLureDeltaMedian}{.000}        
\newcommand{\valLureDeltaLo}{-.110}           
\newcommand{\valLureDeltaHi}{-.066}           
\newcommand{\valLureNNeg}{56}                 
\newcommand{\valLureSignP}{< .001}            
\newcommand{\valLureWilcoxonP}{< .001}        
\newcommand{\valAccTie}{46}                   
\newcommand{\valLureTie}{57}                  
\newcommand{\valHumanAccFirst}{.335}          
\newcommand{\valHumanAccFirstEight}{2.68}     
\newcommand{\valHumanAccFirstSE}{.005}        
\newcommand{\valHumanAccSecond}{.490}         
\newcommand{\valHumanAccSecondEight}{3.92}    
\newcommand{\valHumanAccSecondSE}{.005}       
\newcommand{\valHumanAccDelta}{+.154}         
\newcommand{\valHumanLureFirst}{.458}         
\newcommand{\valHumanLureFirstSE}{.004}       
\newcommand{\valHumanLureSecond}{.146}        
\newcommand{\valHumanLureSecondSE}{.003}      
\newcommand{\valHumanLureDelta}{-.312}        
\newcommand{\valHumanAccFirstPub}{2.7}        
\newcommand{\valHumanAccSecondPub}{3.9}       
\newcommand{\valHumanMaxDev}{0.005}           
\newcommand{\valScaleN}{113}                  
\newcommand{\valScaleAccR}{-.721}             
\newcommand{\valScaleAccP}{< .001}            
\newcommand{\valScaleAccRho}{-.808}           
\newcommand{\valScaleAccRhoP}{< .001}         
\newcommand{\valScaleLureR}{+.652}            
\newcommand{\valScaleLureP}{< .001}           
\newcommand{\valScaleLureRho}{+.718}          
\newcommand{\valScaleLureRhoP}{< .001}        
\newcommand{\valBaseAccR}{+.734}              
\newcommand{\valBaseAccP}{< .001}             
\newcommand{\valConsIntMean}{.946}            
\newcommand{\valConsIntSD}{.081}              
\newcommand{\valConsReaMean}{.990}            
\newcommand{\valConsReaSD}{.045}              
\newcommand{\valConsShift}{+.044}             
\newcommand{\valConsPerfectInt}{.504}         
\newcommand{\valConsPerfectRea}{.903}         
\newcommand{\valItemOneCorrect}{15}           
\newcommand{\valItemOneLure}{30}              
\newcommand{\valItemOneSol}{.26}              
\newcommand{\valItemOneMCE}{.67}              
\newcommand{\valItemOneAccInt}{.812}          
\newcommand{\valItemOneAccRea}{1.000}         
\newcommand{\valItemOneAccDelta}{+.188}       
\newcommand{\valItemOneLureInt}{.087}         
\newcommand{\valItemOneLureRea}{.000}         
\newcommand{\valItemOneLureDelta}{-.087}      
\newcommand{\valItemTwoCorrect}{72}           
\newcommand{\valItemTwoLure}{14}              
\newcommand{\valItemTwoSol}{.40}              
\newcommand{\valItemTwoMCE}{.54}              
\newcommand{\valItemTwoAccInt}{.520}          
\newcommand{\valItemTwoAccRea}{.995}          
\newcommand{\valItemTwoAccDelta}{+.474}       
\newcommand{\valItemTwoLureInt}{.319}         
\newcommand{\valItemTwoLureRea}{.002}         
\newcommand{\valItemTwoLureDelta}{-.317}      
\newcommand{\valItemThreeCorrect}{11}         
\newcommand{\valItemThreeLure}{6}             
\newcommand{\valItemThreeSol}{.55}            
\newcommand{\valItemThreeMCE}{.29}            
\newcommand{\valItemThreeAccInt}{.897}        
\newcommand{\valItemThreeAccRea}{.995}        
\newcommand{\valItemThreeAccDelta}{+.097}     
\newcommand{\valItemThreeLureInt}{.034}       
\newcommand{\valItemThreeLureRea}{.000}       
\newcommand{\valItemThreeLureDelta}{-.034}    
\newcommand{\valItemFourCorrect}{125}         
\newcommand{\valItemFourLure}{120}            
\newcommand{\valItemFourSol}{.37}             
\newcommand{\valItemFourMCE}{.38}             
\newcommand{\valItemFourAccInt}{.924}         
\newcommand{\valItemFourAccRea}{.988}         
\newcommand{\valItemFourAccDelta}{+.064}      
\newcommand{\valItemFourLureInt}{.018}        
\newcommand{\valItemFourLureRea}{.000}        
\newcommand{\valItemFourLureDelta}{-.018}     
\newcommand{\valItemFiveCorrect}{29}          
\newcommand{\valItemFiveLure}{30}             
\newcommand{\valItemFiveSol}{.29}             
\newcommand{\valItemFiveMCE}{.34}             
\newcommand{\valItemFiveAccInt}{.880}         
\newcommand{\valItemFiveAccRea}{.972}         
\newcommand{\valItemFiveAccDelta}{+.092}      
\newcommand{\valItemFiveLureInt}{.064}        
\newcommand{\valItemFiveLureRea}{.004}        
\newcommand{\valItemFiveLureDelta}{-.060}     
\newcommand{\valItemSixCorrect}{30}           
\newcommand{\valItemSixLure}{40}              
\newcommand{\valItemSixSol}{.36}              
\newcommand{\valItemSixMCE}{.59}              
\newcommand{\valItemSixAccInt}{.936}          
\newcommand{\valItemSixAccRea}{.965}          
\newcommand{\valItemSixAccDelta}{+.028}       
\newcommand{\valItemSixLureInt}{.021}         
\newcommand{\valItemSixLureRea}{.019}         
\newcommand{\valItemSixLureDelta}{-.002}      
\newcommand{\valItemSevenCorrect}{4}          
\newcommand{\valItemSevenLure}{9}             
\newcommand{\valItemSevenSol}{.31}            
\newcommand{\valItemSevenMCE}{.19}            
\newcommand{\valItemSevenAccInt}{.922}        
\newcommand{\valItemSevenAccRea}{.982}        
\newcommand{\valItemSevenAccDelta}{+.060}     
\newcommand{\valItemSevenLureInt}{.002}       
\newcommand{\valItemSevenLureRea}{.000}       
\newcommand{\valItemSevenLureDelta}{-.002}    
\newcommand{\valItemEightCorrect}{15}         
\newcommand{\valItemEightLure}{20}            
\newcommand{\valItemEightSol}{.14}            
\newcommand{\valItemEightMCE}{.66}            
\newcommand{\valItemEightAccInt}{.763}        
\newcommand{\valItemEightAccRea}{.996}        
\newcommand{\valItemEightAccDelta}{+.234}     
\newcommand{\valItemEightLureInt}{.179}       
\newcommand{\valItemEightLureRea}{.000}       
\newcommand{\valItemEightLureDelta}{-.179}    
\newcommand{\valItemAccReaMin}{.965}          
\newcommand{\valItemAccReaMax}{1.000}         
\newcommand{\valLureModalItems}{6}            
\newcommand{\valLureShareErrors}{.54}         
\newcommand{\valAccAbove}{67}                 
\newcommand{\valLureBelow}{56}                
\newcommand{\valNCeilingBoth}{46}             
\newcommand{\valNZeroLureBoth}{57}            
\newcommand{\valNZeroLureInt}{57}             
\newcommand{\valNBelowCeiling}{67}            
\newcommand{\valNLureModels}{56}              
\newcommand{\valAccRoomIntMean}{.716}         
\newcommand{\valAccRoomIntLo}{.659}           
\newcommand{\valAccRoomIntHi}{.769}           
\newcommand{\valAccRoomReaMean}{.977}         
\newcommand{\valAccRoomReaLo}{.955}           
\newcommand{\valAccRoomReaHi}{.994}           
\newcommand{\valAccRoomDeltaMean}{+.261}      
\newcommand{\valAccRoomDeltaMedian}{+.200}    
\newcommand{\valAccRoomDeltaLo}{+.216}        
\newcommand{\valAccRoomDeltaHi}{+.308}        
\newcommand{\valAccRoomNPos}{67}

\newcommand{\valClosedMean}{.963}             
\newcommand{\valClosedLo}{.933}               
\newcommand{\valClosedHi}{.987}               
          
\newcommand{\valClosedFull}{56}               
\newcommand{\valLureRoomIntMean}{.182}        
\newcommand{\valLureRoomIntLo}{.152}          
\newcommand{\valLureRoomIntHi}{.215}          
\newcommand{\valLureRoomReaMean}{.006}        
\newcommand{\valLureRoomReaLo}{.002}          
\newcommand{\valLureRoomReaHi}{.012}          
\newcommand{\valLureRoomDeltaMean}{-.176}     
\newcommand{\valLureRoomDeltaLo}{-.207}       
\newcommand{\valLureRoomDeltaHi}{-.147}       
              
\newcommand{\valLureRoomNNeg}{56}             
\newcommand{\valLureRoomSignP}{< .001}        
                
\newcommand{\valElimMean}{.969}               
\newcommand{\valElimLo}{.937}                 
\newcommand{\valElimHi}{.992}                 
            
\newcommand{\valElimFull}{49}                 
\newcommand{\valAccRoomScaleR}{-.585}         
\newcommand{\valAccRoomScaleP}{< .001}        
\newcommand{\valLureRoomScaleR}{+.463}        
\newcommand{\valLureRoomScaleP}{< .001}       
\newcommand{\valClosedScaleRho}{+.521}        
\newcommand{\valClosedScaleRhoP}{< .001}      
\newcommand{\valClosedScaleR}{+.462}          
\newcommand{\valClosedScaleP}{< .001}         
\newcommand{\valElimScaleRho}{+.466}          
\newcommand{\valElimScaleP}{< .001}           
\newcommand{\valElimScaleR}{+.356}            
\newcommand{\valElimScalePearP}{= .007}       
\newcommand{\valNonLureIntMean}{.078}         

\newcommand{\valNonLureReaMean}{.010}         

\newcommand{\valRelLure}{.97}                 
\newcommand{\valRelNonLure}{.87}              
\newcommand{\valHumanRelLure}{.68}            
\newcommand{\valOrgN}{39}                     
\newcommand{\valOrgAccPos}{30}                
\newcommand{\valOrgAccZero}{9}                
\newcommand{\valOrgAccSignP}{< .001}          
\newcommand{\valOrgAccWilcoxonP}{< .001}      
\newcommand{\valOrgLureNeg}{28}               
\newcommand{\valOrgLureZero}{11}              
\newcommand{\valOrgLureSignP}{< .001}         
\newcommand{\valOrgLureWilcoxonP}{< .001}     

\title{Large-Language Models Validate the Cognitive Reflection Test}
\shorttitle{LLMs Validate the CRT}

\authorsnames{Devon Yanitski}
\authorsaffiliations{{Department of Psychology, University of Alberta, Edmonton, Alberta, Canada\\
  \href{mailto:dyanitsk@ualberta.ca}{dyanitsk@ualberta.ca}\\[1.2em]
  \today\\[0.6em]\mbox{}}}

\authornote{%
\addORCIDlink{Devon Yanitski}{0009-0006-1568-3387}

This research did not receive any specific grant from funding agencies in the public, commercial, or not-for-profit sectors. Computing resources were provided by the Digital Research Alliance of Canada. The author has no competing interests to declare.

Correspondence concerning this article should be addressed to Devon Yanitski,
Department of Psychology, University of Alberta, P-217 Biological Sciences Building,
Edmonton, Alberta T6G 2E9, Canada. Telephone: 587-280-9838. E-mail: dyanitsk@ualberta.ca}

\shortcites{meyer-etal-2024, deneys-etal-2013, travers-etal-2016, szaszi-etal-2017,
  toplak-etal-2011, toplak-etal-2014, pennycook-etal-2015, pennycook-etal-2016,
  baron-etal-2015, liberali-etal-2012, welsh-etal-2013, otero-etal-2022,
  patel-etal-2019, sirota-etal-2021, meyer-etal-2018, agarwal-etal-2025,
  aher-etal-2023, deng-etal-2024, dillion-etal-2023, dong-etal-2024,
  ghosal-etal-2025, hagendorff-etal-2023a, hagendorff-etal-2023b, kojima-etal-2022,
  ma-etal-2025, moskovitz-etal-2024, nye-etal-2021, pennycook-etal-2015b,
  shaki-etal-2023, thompson-etal-2011, wei-etal-2022, yang-etal-2023, yax-etal-2024,
  kwon-etal-2023, fedus-etal-2022, gholami-etal-2022}

\abstract{%
The Cognitive Reflection Test (CRT) measures the disposition to override a compelling
but wrong intuitive answer, yet in humans reflection is a latent trait: no design can
assign a person to be reflective, so numeracy, motivation, and prior exposure entangle
every score. We administered the updated eight-item CRT \citep{meyer-etal-2024} to
\valNComplete{} open-weight large language models (LLMs) under two conditions: \emph{intuition}, in which the model
answered in a single pass with reasoning switched off, and
\emph{reason}, in which the same model first generates reasoning tokens. We excluded \valNAware{} models showing textual
signs of recognizing the items. Across
the \valNAnalyzed{} analyzed models, reasoning raised mean accuracy from \valAccIntMean{} to \valAccReaMean{} and
cut the intuitive-lure rate (the published most common error) from \valLureIntMean{} to \valLureReaMean{},
reproducing in direction the human first-exposure-to-second-chance effect \citep{meyer-etal-2024} (accuracy \valHumanAccFirst{} to \valHumanAccSecond{}, lure \valHumanLureFirst{} to \valHumanLureSecond{}, recomputed from their data), without the error
feedback the human procedure provides. The effect never reversed across \valNOrgs{}
organizations, several reasoning implementations, and widely varying model sizes.
A ceiling effect appeared in \valNCeilingBoth{} models (perfect in both conditions), suggesting that, on these items, many LLMs' non-deliberative responses already exceed human reflective performance. Among
the \valNBelowCeiling{} below-ceiling models, the accuracy gain was $\valAccRoomDeltaMean$. The replication converts
cognitive reflection from a latent between-person trait into a
within-agent manipulation (with added computation), suggests dual-process organization is not specific to the human substrate, and
motivates the hypothesis that human System~2 is System~1 iterated through a serial,
language-based loop.
}

\begin{document}
\maketitle



The Cognitive Reflection Test (CRT) is built on a simple behavioral premise: pose a
question that reliably summons a specific wrong answer, and observe whether the
respondent reports that answer or stops to check it. In the canonical item (``A bat
and a ball cost \$1.10 in total. The bat costs \$1.00 more than the ball. How much
does the ball cost?''), the response that springs to mind, 10 cents, is wrong upon
further reflection. \citet{frederick-2005} defined the trait the test targets,
cognitive reflection, as the ability or disposition to resist reporting the response
that first comes to mind. Across more than 3,000 respondents, scores on three such
items varied widely and predicted consequential behavior, including measures of time
preference and risk preference \citep{frederick-2005}.

The test's logic derives from the account of heuristic judgment developed by
\citet{kahneman-frederick-2002, kahneman-frederick-2005}, in which a fast intuitive
process proposes answers by attribute substitution (answering an easier, more
accessible question in place of the one that was asked) while a slower supervisory
process endorses, corrects, or overrides those proposals. Because that monitoring is
normally quite lax, fluent but mistaken intuitions often pass uncorrected
\citep{kahneman-frederick-2002}. A CRT item operationalizes this architecture in a
single response: substitution yields a determinate, attractive, and incorrect answer,
so the answer given reveals whether the intuition was endorsed or overridden.

What distinguishes the CRT from an ordinary difficult test is therefore the structure
of its errors. Each item has a dominant intuitive lure, a single most common error
that accounts for a large share of incorrect responses \citep{meyer-etal-2024}, so a wrong answer signals an
intuition left unchecked, not an inability to solve the problem: in able samples, most
respondents who miss these problems are careless rather than hopeless (able to
compute the correct answer but failing to check the one that came to mind), though
the careless-to-hopeless balance shifts with the population's cognitive ability
\citep{meyer-frederick-2023}. Process
evidence supports this reading. Respondents who give the lure nonetheless register the
conflict, reporting lower confidence than on matched control versions of the problems
that lack the lure \citep{deneys-etal-2013}. Mouse-tracking shows answers gravitating
toward the lure early in a trial, even when the correct response ultimately wins out
\citep{travers-etal-2016}. Think-aloud protocols reveal heterogeneous solution
paths in which reflection neither guarantees a correct answer nor is always needed for
one \citep{szaszi-etal-2017}. Together, these findings motivate scoring the CRT on two
complementary measures: overall accuracy and, separately, the rate of lure responses,
the signature of an intuition reported unchecked.

\subsection{The CRT Measures a Reflection Trait Distinct From Numeracy}

Much of the CRT's influence rests on its predictive record. The test predicts
performance on batteries of heuristics-and-biases tasks even after measures of
intelligence, executive functioning, and thinking dispositions are partialed out
\citep{toplak-etal-2011}, a result that holds, with cognitive ability and thinking
dispositions controlled, for an expanded seven-item version built
in part to outrun the original items' growing fame \citep{toplak-etal-2014}. Lower
scores accompany greater receptivity to pseudo-profound bullshit
\citep{pennycook-etal-2015}. Whether lure responses are additionally informative in
their own right (indexing reliance on intuition rather than merely the absence of
reflection) is itself contested: \citet{pennycook-etal-2016} tested exactly this
and concluded that the CRT validly measures reflective but not intuitive thinking. Our
use of the lure rate assumes less than that debate contests: it tracks each item's
engineered, published most common error as the signature of an unchecked intuition,
not as a trait index of intuitiveness.

That predictive record did not settle what the test measures. Because the classic
items are arithmetic word problems, CRT scores
correlate strongly with numeracy, and several lines of evidence suggested the test
might be little more than a math test: correlational and factor-analytic evidence attributed most of its
predictive power to numerical ability \citep{welsh-etal-2013}, factor analyses found
CRT and numeracy items co-loading on a shared dimension in one of two samples
\citep{liberali-etal-2012}, a
meta-analysis estimated large corrected correlations with cognitive abilities,
numerical ability prominent among them \citep{otero-etal-2022}, and the lure-bearing
items proved psychometrically ordinary, indistinguishable by item-quality measures
from matched arithmetic items \citep{attali-bar-hillel-2020}. Decomposing responses into lure
detection and arithmetic suggested that calculation carried much of the test's
prediction of decision quality \citep{sinayev-peters-2015}, though a commentary countered
that the items' numerical format may itself shape performance on both the CRT and
numeracy tests \citep{mastrogiorgio-2015},
and CRT performance also tracks insight problem solving and numeracy in ways that
complicate a pure reflection reading \citep{patel-etal-2019}. Even the lures'
interpretation was challenged: comparable predictive power emerges from matched items
without lures \citep{baron-etal-2015}, and giving the lure can be defended as rational
deference to the answer a question's wording implicitly recommends, rather than as
miserliness \citep{levy-2023}. On the other side,
mathematical modeling indicated that the CRT measures reflection over and above
numeracy \citep{campitelli-gerrans-2014}, and a validated verbal CRT showed that
lure-based reflection items need not involve arithmetic at all \citep{sirota-etal-2021}.

An adversarial collaboration among the test's proponents and critics went far toward
settling the dispute \citep{meyer-etal-2024}. Pitting lure-bearing items against comparably difficult
mathematical items without lures, structural equation modeling recovered a reflection
factor distinct from mathematical ability, and a dual-response paradigm, in which
respondents are told which items they missed and given a second chance to revise them,
showed scores improving from first exposure to second chance and the second-chance score
behaving more like an ordinary math test, as expected if first-exposure errors reflect
unchecked intuitions rather than missing mathematical ability. The
collaboration concluded that cognitive reflection is a distinct and measurable trait
(while noting that its strongest claims rest on corrections for measurement error),
and it published an updated eight-item CRT, each item with a documented most
common error \citep{meyer-etal-2024}. That instrument, its published lures, and its
first-exposure versus second-chance benchmark are the foundation of the present study.

\subsection{Familiar Items Undermine What a Score Means}

A further threat to the classic test was its own celebrity. Among online volunteers,
over half reported having previously seen at least one of the original items, and
previously exposed respondents scored markedly higher \citep{haigh-2016}. In a large
age-stratified community sample, 44\% reported prior experience with the test or
similar tasks, again with substantially higher scores, prompting a recommendation to
adopt newer multi-item versions \citep{stieger-reips-2016}. Item familiarity was an
explicit motivation for successor instruments, from the expanded CRT
\citep{toplak-etal-2014} to the verbal CRT \citep{sirota-etal-2021}. For human respondents the threat should not be
overstated: tracking 14,053 individuals across up to 25 administrations,
\citet{meyer-etal-2018} found that each exposure raised scores by only 0.024 items on
average and left predictive validity intact, with self-reported exposure exaggerating
the effect. But the same work noted that worked solutions to the classic items are
abundant online, including hundreds of explainer videos for the bat-and-ball problem
alone \citep{meyer-etal-2018}.

For large language models (LLMs) trained on web-scale corpora, that abundance converts
a manageable nuisance into a central interpretive problem: the classic items and their
solutions are likely present in training text \citep{hagendorff-etal-2023b}, so a correct answer could
reflect retrieval of a memorized solution rather than anything resembling reflection.
Benchmark leakage of exactly this kind is a documented risk in LLM
evaluation \citep{deng-etal-2024}, inflating scores in ways that survive even
paraphrase \citep{dong-etal-2024, yang-etal-2023} or translation \citep{yang-etal-2023} of
the leaked items. We addressed it in
three ways. First, we administered the updated eight items of
\citet{meyer-etal-2024}, an instrument published only in late 2024, in the hope that
its items are rare in the training data of the models we tested. Three of its items are new
variants of the classic three, and two others recast items of an earlier expanded CRT in
new surface content (see Method), so this first defense is partial.
Second, as described in the Method, we screened every response in an awareness scan
for signs that a model recognizes the test, its items, or its source, and we excluded
flagged models. Third, the lure rate itself functions as a check on memorization: a
model reading answers off a memorized key would simply recite them. It would not, with
reasoning off, reproduce the very errors that trap people.

Familiarity is only about whether the answers can be recalled. What the errors
\emph{mean} (and what correcting them implies) comes from the dual-process theory
behind the test, which we turn to next.

\subsection{Reasoning Models Make Deliberation a Switch}

The disposition the CRT is designed to elicit (overriding a compelling but wrong
intuition) sits at the center of dual-process theories of human reasoning. On the default-interventionist account that \citet{evans-stanovich-2013}
defend, a fast, autonomous Type~1 process delivers a default answer, which a slower,
effortful Type~2 process may or may not intervene upon and revise. What distinguishes
the two, in their analysis, is not speed or accuracy per se but cognitive decoupling and
a heavy load on working memory: Type~2 thinking sustains hypothetical, serial
computation, whereas Type~1 runs autonomously and cheaply. A CRT item exploits exactly
this architecture: the dominant lure described above is the Type~1 default, and a
correct answer requires detecting that this default conflicts with a more considered
solution and then overriding it.

The detection step is itself a documented component of the mechanism. In base-rate
problems, for example, people register the conflict between a tempting intuition and a
competing logical response even when they ultimately go with the intuition, an early,
largely automatic signal that more deliberation may be warranted
\citep{pennycook-etal-2015b}. \citet{bago-deneys-2017} push this further, showing with a
two-response procedure (an immediate, speeded first answer followed by a
free-time second answer) that a correct, logical response is often already present in
the fast initial response, so that deliberation frequently confirms or strengthens an
intuition rather than building one from scratch. The decision of whether to engage that
second, deliberative pass is governed metacognitively: \citet{thompson-etal-2011} find
that a low ``feeling of rightness'' attached to the initial answer predicts longer
rethinking and a greater chance of revising it. This two-response logic, the same
person answering once quickly and once with deliberation, is the human template the
present study adapts.

The serial, effortful pass that override requires is, on the standard view, a function
of working memory. \citet{baddeley-hitch-1974} provided early evidence for this by loading reasoning,
comprehension, and recall with a concurrent memory task and observing
interference once the load approached memory span, evidence for a limited-capacity workspace shared between storage and
processing. The resulting model couples a central executive to subsidiary stores,
including a phonological loop that maintains speech-based information through subvocal
rehearsal \citep{baddeley-1992}. That loop is widely read as the substrate of inner
speech, an interplay between an ``inner voice'' that generates and an ``inner ear'' that
re-reads its own output \citep{buchsbaum-2013}, and inner speech is in turn argued to be
not merely a way of reporting thought but a medium of thinking itself, the linguistic
scaffold on which deliberation and problem-solving are carried out
\citep{kompa-2024}. Deliberation, on this composite picture, is the recursive use of a
language-based working-memory buffer to run serial steps that a single intuitive pass
cannot. Crucially, the dual-process organization need not be a quirk of human
neurobiology: \citet{moskovitz-etal-2024} give it a normative, computational account in
which a simple, compressed controller and a more complex one that overrides it emerge
from a single learning objective, implemented as two neural networks. That framing
invites the question of whether the same fast--slow dissociation appears in other systems
that compute over language.

LLMs offer one such system, and they instantiate a strikingly literal version of the
fast--slow distinction. A transformer answering in a single forward pass
emits its response through one wave of propagation, with no opportunity to revisit its
starting assumptions (a fast, autonomous pattern completion). Deliberation is added by
letting the model first generate intermediate tokens that are written into its own
context and re-ingested before it commits to an answer. \citet{wei-etal-2022} showed
that prompting a model to produce such a chain of thought sharply improves performance
on multi-step arithmetic and reasoning problems, and \citet{kojima-etal-2022} showed
that a single instruction (``let's think step by step'') elicits a similar,
if smaller, gain without worked examples. \citet{nye-etal-2021} had earlier framed the same idea as a
``scratchpad'' that supplies the model with adaptive computation time and an external
buffer for intermediate state. The reason this matters is not merely empirical:
\citet{merrill-sabharwal-2024} prove that intermediate generation expands the
computational power of a transformer, with each additional decoding step buying
computational power beyond the fixed limit of one forward pass. The intermediate-token loop is thus, formally, a way of trading serial time
for computational depth, a precise mechanical counterpart to the working-memory
account of human deliberation.

This parallel has made the cognitive-science toolkit a natural instrument for probing
models. A growing body of work administers tasks built for humans to LLMs and reads
their outputs as behavior, treating the model as a participant
\citep{aher-etal-2023, dillion-etal-2023, shaki-etal-2023}. The same lens has been turned on
reasoning errors and cognitive biases \citep{jones-steinhardt-2022,
macmillan-scott-musolesi-2024} and, with comparison against fresh human samples, on
bounded rationality itself \citep{yax-etal-2024}. The most direct precedent applies the
CRT and related illusions across a model family in explicit System~1/System~2 terms:
\citet{hagendorff-etal-2023b} report that the larger GPT-3-era models fall for the
intuitive lures even more often than people do (a tendency the most capable models in their
study had learned to avoid), that chain-of-thought reasoning helps models escape the
traps, and that the context window can serve as an external short-term memory, a
notepad for System-2-like processing \citep[see also][]{hagendorff-etal-2023a}. These studies
caution, however, that the analogy is imperfect: models can pattern-match to
training-like surface forms rather than reason, so that small structural changes to a
problem collapse their accuracy \citep{ma-etal-2025}, and their errors do not always
mirror the human ones \citep{macmillan-scott-musolesi-2024}.

Two qualifications shape how we build on this work. First, prompt-elicited chain of
thought is the conceptual ancestor of, not the same mechanism as, the deliberation our
study manipulates. The earliest demonstrations engineered reasoning by inserting
exemplars or instructions into the prompt \citep{wei-etal-2022, kojima-etal-2022}.
Recent models instead expose a trained-in reasoning mode that a documented chat-template
control switches on or off \citep{qwen-team-2025}, so that deliberation can be toggled while the task text is
held byte-for-byte constant. Second, the amount of deliberation is a graded quantity, not a
binary one: performance scales with test-time computation, though not without
limit, and longer reasoning traces can even degrade accuracy through ``overthinking''
\citep{agarwal-etal-2025, ghosal-etal-2025}. A within-model switch between no
deliberation and trained deliberation therefore isolates the cleanest version of the
contrast: whether engaging the serial, token-based loop, at fixed weights, reproduces
the human reflection signature.

\subsection{This Study Runs the CRT Twice Within Each Model}

The present study uses that switch to run the CRT through a single agent twice. We administered the updated eight-item CRT of \citet{meyer-etal-2024} to a
large collection of open-weight language models under two conditions that differ only in the
model's own chat-template reasoning toggle: an \emph{intuition} condition, in which the model emits
its answer directly in a single pass, and a \emph{reason} condition, in which the same
model (the same frozen weights) first generates intermediate reasoning tokens and
only then answers. Because the item text, the sampling parameters, and the weights are
identical across conditions, the manipulation removes everything that would otherwise
vary between respondents. What remains is the engagement of serial, token-based
deliberation, together with the additional test-time computation that generating
reasoning tokens entails (a covariation intrinsic to the manipulation, to which we
return in the Discussion). This is the within-agent analogue of the human
first-exposure-to-second-chance procedure, the same agent answering once with
deliberation off and once with it on \citep{meyer-etal-2024, bago-deneys-2017,
thompson-etal-2011}, but a stricter one, since, unlike the human second chance, the model
receives no feedback about which answers were
wrong. It is merely permitted to think.

We tested two directional hypotheses, framed to match the within-model design. First,
enabling reasoning should raise CRT accuracy: the deliberative pass should let the model
override the intuitive answer it would otherwise give. Second, and more diagnostically,
enabling reasoning should specifically lower the rate at which models produce the
dominant intuitive lure (the engineered most-common error) rather than merely
reducing errors in general, reproducing the mechanism the test was built around rather
than its outcome alone. We expected these effects to hold across a heterogeneous
population of model families and across the several distinct implementations of the
reasoning toggle. To the extent that reasoning in LLMs
reproduces the direction of the human first-exposure-to-second-chance
effect that \citet{meyer-etal-2024} report, the replication does more than support the
claim that reflection is separable from baseline mathematical competence: it converts a latent human trait into a
manipulable, within-agent cause, and turns a system whose mechanism is fully specified
into a starting point for hypotheses about how human deliberation is organized.

\newpage
\section{Method}

\subsection{Models}

The study population was a collection of open-weight LLMs, each
served from its published weights. A model is open-weight when its trained parameters (its
weights) are published for download, so that anyone can run it on their own hardware.
Eligibility required that a model support both response modes of the experimental
manipulation. It had to be able to emit a final answer directly, without intermediate
reasoning tokens, \emph{and} to generate an extended chain of intermediate reasoning tokens
before answering, with the difference controlled by the model's own switch (the
\emph{reasoning toggle}, see the next subsection). Models whose chat templates
unconditionally enable reasoning, with no mechanism to disable it, could not realize the
intuition condition and were excluded from the study.

The sample is the one assembled for a companion study of moral judgment
\citep{yanitski-2026}. Those runs also presented the eight CRT items (see Procedure), and
the CRT trials are archived with the study's data \citep{yanitski-2026-crt-data}.
The sample comprises the \valNComplete{} models whose reasoning toggle passed the two checks described in
the next subsection, an audit of the rendered prompts and a manipulation check. Of these,
\valNNewRun{} were collected in new runs for the two studies. The other \valNReused{}
contribute CRT trials from an earlier collection on the same infrastructure with the same
toggle and the same prompts (see Procedure).

Of the \valNComplete{} models, \valNAware{} were excluded by the awareness scan (see the
Awareness Scan subsection and Results), leaving \valNAnalyzed{} models in every analysis.
The \valNAnalyzed{} analyzed models come from \valNOrgs{} developing organizations:
\valOrgList{}. They span more than three orders of magnitude in scale, from
\valParamMin{} to \valParamMax{} billion total parameters (median \valParamMedian{}
billion), and include both dense and mixture-of-experts architectures. In a dense model,
every parameter takes part in producing every token. In a mixture-of-experts model, each
layer holds several alternative sub-networks (the experts), and each token is routed through
only a few of them \citep{fedus-etal-2022}, so we report total parameters throughout. Because every model was run
from fixed published weights, the ``participant'' in each paired comparison is a frozen
network: nothing about the model changes between conditions. The full model roster appears
in the Supplementary Materials.

Every model ran from downloaded weights on Digital Research Alliance of Canada computing
clusters, and no model was accessed through a commercial provider or any other hosted
service. Of the \valNNewRunAn{} analyzed models from the new runs, \valNvllmNew{} were served
on graphics processing units (GPUs) by vLLM \citep{vllm-2026, kwon-etal-2023}, an inference
server program, and \valNllamaNew{} were served on central processing unit (CPU) nodes by
llama-server, the server program of the llama.cpp inference library \citep{llamacpp-2026}.
llama-server reads model weights from GGUF files, the model file format of that library,
which here held quantized weights. Quantization refers to storing a model's weights at a
lower numerical precision than the one they were trained at, so that a large model fits in
the available memory \citep{gholami-etal-2022}. These files held 4-bit weights for most of the llama.cpp models and
8-bit weights for a few of the smallest. Some of the vLLM models were likewise served from
quantized checkpoints, at 8-bit or 4-bit precision, a few of them released by third parties
rather than by the developer. The earlier collection used the same two servers.
The per-model serving settings, including any quantization, and the total parameter counts
used in the scaling analyses are listed in the project repository (see the Open Practices
Statement).

\subsection{Reasoning Toggle}

Every model answered every item under two conditions that differ only in whether the
model's reasoning mode is engaged. In the \emph{intuition} condition, reasoning is
disabled: the model emits its answer directly, in a single pass, normally without
producing intermediate reasoning tokens (the manipulation check below bounds how often it
did). In the \emph{reason} condition, reasoning is enabled. The
same model, with the same weights, first generates intermediate ``thinking'' tokens, which
are written into its own context and re-read autoregressively, and only then produces the
final answer. The toggle is thus the within-model analogue of the contrast between an
immediate first response and a deliberate second pass
\citep{bago-deneys-2017, thompson-etal-2011}. The same agent answers twice, once with
deliberation off and once with deliberation on, at fixed weights.

Open-weight models do not share a single switch, so the toggle was realized through each
model's own documented control mechanism, audited against the model's chat template. A
chat template refers to the model's own formatting code, shipped with its weights, that
turns a conversation into the exact token sequence the model was trained on
\citep[see, e.g.,][]{qwen-team-2025}. That sequence
includes control markup, the special tokens or tags that the model reads as formatting
rather than as content, such as the \texttt{<think>} and \texttt{</think>} tags that open
and close a reasoning block. We required the toggle to be an argument to the chat
template. With the argument on, the template ends the prompt so that the model begins a
reasoning block, and with it off, the template closes or omits that block. The argument is
named \texttt{enable\_thinking} for \valNkeyEnableThinking{} of the analyzed models and
\texttt{thinking} for \valNkeyThinking{}, and the remaining \valNkeyOther{} use another
name, such as \texttt{reasoning\_effort} or \texttt{thinking\_mode} (see the Supplementary
Materials). Because the model's own formatting code, not text written by us, does the
switching, the message we send can be identical in both conditions. Models whose toggle
works by adding text to the conversation, such as a control token in the user turn or a
directive in a system message, were not eligible. An instruction to think, or not to
think, is itself content that could change the answer, so such a toggle would confound the
reasoning manipulation with a change in the prompt.

Two checks confirmed that each toggle did what it was meant to do. First, for every model
we rendered the same user message through its chat template with the toggle on and with
it off and compared the two token sequences. The two renderings had to differ only in
control markup, which was the case for \valNtrueOnlyAn{} of the analyzed models. For the
other \valNsettingsAn{}, the renderings also differed in a single line that the template
itself writes to state a reasoning setting, such as the effort level, and we allowed that
difference. Any difference in natural-language instruction text excluded the model.
Second, each model had to pass a manipulation check, because some models ignore the
toggle and reason in neither condition or in both. A trial counted as containing reasoning
when more than 20 characters of reasoning text remained after the control markup and any
copy of the requested answer were removed. A model passed when reasoning was present in at
least 80\% of its reason-condition trials and in at most 20\% of its intuition-condition
trials. Its reason-condition trials also had to use, on average, at least twice as many
completion tokens as its intuition-condition trials. Completion tokens are the tokens the
model generates in its reply, reasoning included. All three quantities were computed over
all 56 items of the model's run (for the reused models, the same items from their earlier
runs), which were the eight CRT items, the 40 dilemmas of the
companion study \citep{yanitski-2026}, and the eight items of a third task (see
Procedure). The earlier collection sometimes stored no separate reasoning field, so for
the reused models a reply of 50 or more completion tokens also counted as containing
reasoning (a bare answer runs to about 7 to 15 tokens). The check excluded \valNexclManip{}
models whose toggle was otherwise eligible. Among the analyzed models, reasoning was
present in at least \valManipReasonMin\% of reason-condition trials and in at most
\valManipIntuMax\% of intuition-condition trials, and the smallest ratio of
reason-condition to intuition-condition completion tokens was \valManipRatioMin. On the
CRT items alone, the median model wrote \valTokIntMedian{} completion tokens per trial
under intuition and \valTokReaMedian{} under reason.

The task input was held constant across conditions. The user message (the item's
vignette, its question, and the response instruction) was byte-identical under intuition
and reason, and no system message was sent in either condition. Several model cards
recommend different sampling settings (temperature, top-$p$, top-$k$, and any
model-specific penalties) for thinking and non-thinking use, and developers evaluate the
two modes with different settings \citep[e.g.,][]{qwen-team-2025}. We pinned the sampling
settings at each model's recommended thinking-mode values in both conditions, because a
change in the sampling distribution would otherwise be confounded with the reasoning
manipulation. The two conditions therefore differ only in the toggle and the serial
computation it engages, not in prompt content or sampling distribution. The exact
per-model settings are listed in the project repository (see the Open Practices
Statement).

\subsection{Stimuli}

The stimuli were the eight items of the updated CRT introduced by
\citet{meyer-etal-2024}, transcribed from their Table~1. The set comprises
variants of the three original CRT items \citep{frederick-2005} together with five other
items that its authors expected to function in the same way. Two of these (items 5 and 7)
recast items of an earlier expanded CRT \citep{toplak-etal-2014} in new surface content,
with the same correct answers and lures. Each item admits a single correct
numeric answer and a dominant intuitive lure. The lure is the published most common error
(MCE), the answer produced by performing a simple but inappropriate operation on the surface
form of the problem \citep{meyer-etal-2024}. The updated items were chosen over the original three because the
instrument is recent and its items are far less culturally ubiquitous than the classic
bat-and-ball problem \citep{haigh-2016, stieger-reips-2016}, and were therefore less
likely to be heavily represented in the models' training corpora
\citep{hagendorff-etal-2023b} (a risk we additionally probed directly, see the Awareness
Scan subsection). Table~\ref{tab:crt-items} lists the eight items as administered,
together with the correct answer, the intuitive lure, and the published human solution
and MCE rates.

The only changes from the published format concerned the response format. Where
\citet{meyer-etal-2024} print fill-in-the-blank response fields (e.g., ``How much does
the hammer cost? \$\_\_\_''), the blank was rendered as an explicit question with the
unit named in the question text (e.g., ``How much does the hammer cost, in dollars?''),
so that a bare number constitutes a complete answer. For the same reason, items 3 and 6,
which end in a sentence to be completed, were rephrased as direct questions. Item content
was otherwise unchanged.

\begin{table}[htbp]
\caption{The Eight Updated CRT Items, With Correct Answers, Intuitive Lures, and Published Human Response Rates}
\label{tab:crt-items}
\centering
{\renewcommand{\baselinestretch}{1}\footnotesize
\begin{tabular}{@{}c p{8.4cm} r r r r@{}}
\toprule
Item & Text as administered & Correct & Lure & Sol.\ rate & MCE rate \\
\midrule
1 & A drill and a hammer cost \$330 in total. The drill costs \$300 more than the hammer. How much does the hammer cost, in dollars? & \valItemOneCorrect{} & \valItemOneLure{} & \valItemOneSol{} & \valItemOneMCE{} \\
2 & A dog and a cat weigh 100 pounds in total. The dog weighs 86 pounds. What is the difference in weight between the dog and the cat, in pounds? & \valItemTwoCorrect{} & \valItemTwoLure{} & \valItemTwoSol{} & \valItemTwoMCE{} \\
3 & After hatching from its egg, a baby bird doubles in weight every day. On day 12 it weighs a pound. On what day did the bird weigh half a pound? & \valItemThreeCorrect{} & \valItemThreeLure{} & \valItemThreeSol{} & \valItemThreeMCE{} \\
4 & When it's on sale for 20\% off, a toaster costs \$100. When it's not on sale, how much does it cost, in dollars? & \valItemFourCorrect{} & \valItemFourLure{} & \valItemFourSol{} & \valItemFourMCE{} \\
5 & Rachel is the 15th tallest and the 15th shortest girl in her class. How many girls are in her class? & \valItemFiveCorrect{} & \valItemFiveLure{} & \valItemFiveSol{} & \valItemFiveMCE{} \\
6 & If 30 elves can wrap 30 gifts in 30 minutes, then how many minutes would it take 40 elves to wrap 40 gifts? & \valItemSixCorrect{} & \valItemSixLure{} & \valItemSixSol{} & \valItemSixMCE{} \\
7 & Jack can drink a bottle of cider in 6 days. Jill can drink a bottle of cider in 12 days. How long would it take them, in days, if they were drinking from the same bottle? & \valItemSevenCorrect{} & \valItemSevenLure{} & \valItemSevenSol{} & \valItemSevenMCE{} \\
8 & In a basket of 60 apples that are either green or red, green apples are only 1/3 as common as red apples. How many apples are green? & \valItemEightCorrect{} & \valItemEightLure{} & \valItemEightSol{} & \valItemEightMCE{} \\
\bottomrule
\end{tabular}
}
\caption*{\textit{Note}. Lure = the published most common error (MCE). Sol.\ rate and
MCE rate are the published human first-exposure frequencies of the correct answer and
the MCE, respectively, from Table~1 of \citet{meyer-etal-2024}. Item content is
as published, except that fill-in-the-blank response fields were rendered as explicit
questions with the unit named in the question text.}
\end{table}

\subsection{Design}

The experiment used a $\valNCond \times \valNItems \times \valNReps$ within-model
design: \valNCond{} conditions (intuition vs.\ reason, manipulated by the reasoning
toggle) $\times$ \valNItems{} CRT items $\times$ \valNReps{} repetitions, for
\valNTrialsPerModel{} trials per model. Condition was a within-model factor: every
model answered every item in both conditions, and each combination of item and condition was
sampled \valNReps{} times in independent requests. What is held fixed across the two
conditions is the model's weights, the byte-identical item text, and the sampling
parameters. The only difference is whether the model performs serial test-time
deliberation, generating intermediate reasoning tokens, before answering. This
difference also adds test-time computation, the reasoning tokens themselves, so the
contrast is deliberation-plus-its-computation versus neither (see Boundaries of the
Inference).

This design mirrors, within a single agent, the dual-response procedure that
\citet{meyer-etal-2024} used with human respondents. In their procedure, the same
person answers every item on first exposure and then, after being told which items were
missed, gets an unanticipated second chance at those items. Here, the same model answers
each item twice: once with deliberation off (intuition, the analogue of first exposure)
and once with deliberation on (reason, the analogue of the second chance). In both
designs the agent is constant and only the engagement of deliberation varies. What
differs between them is that the human second chance includes error feedback, because
participants are told which answers were wrong, whereas the reason condition for the
LLMs supplies no feedback of any kind.

\subsection{Procedure}

Each newly collected model was run by an automated runner, a script that drives the whole
collection for one model, on GPU or CPU nodes of five Digital Research Alliance of Canada
clusters (Fir, Rorqual, Nibi, Trillium, and Narval). The runner first started a local
server that loaded the model's weights and accepted the request format of commercial chat
services, so the same runner worked for every model. Before any trial, the runner repeated
the rendered-prompt audit described above against the chat template that the server was
actually using, and it stopped unless the two conditions differed only as that audit
allows. It then issued the \valNTrialsPerModel{} CRT trials, one request per trial,
interleaved in a shuffled order with the 400 dilemma trials of the companion study
\citep{yanitski-2026} and 80 trials of an eight-item third task. Each reply was limited to
24,576 tokens.

The prompt for every trial consisted of the item's vignette, its question, and a fixed
response instruction, joined by blank lines. The response instruction was identical for
all items and both conditions: ``Respond with ONLY a JSON object of the form
\{\texttt{"answer"}: N\} where N is the numeric answer to the problem (a number only,
with no units or other text).'' Each request held this single user message, with the
toggle set for its condition.

The runner stored every reply in full, split into the reasoning trace and the answer text
(see Scoring), together with its token counts and the reason the reply ended. Every
attempt, valid or not, was appended as one row to an append-only JSON-lines file per model,
and rows were never edited or overwritten. A trial was retried, up to three attempts in
all, when the request failed, when its reply yielded no answer (including an empty reply
and a reply cut off at the output limit), or when the reasoning block the template opened
was never closed. On restart, the runner reconstructed from its file which trials still
needed work, so interrupted collections continued without duplicating completed trials.

The other \valNReused{} models were not collected again. Their CRT trials come from an
earlier collection that used the same toggle and the same user message, with the models
served by vLLM or llama-server on the same clusters. That collection differed in three
operational respects. The CRT trials were not interleaved with other tasks. The output
limit was the smaller of the model's configured maximum and a per-run cap of 4,096 tokens,
which was raised for models with long reasoning traces. A trial was retried at most once,
so up to two attempts. That collection also did not record whether a reply ended at its
limit. A model's replies were therefore treated as cut off if at least three of them
reached the same maximum length of 4,000 tokens or more, the signature of a fixed limit,
and no reused trial met this criterion. The stored replies of the reused models were
scored by the same rule as the new runs (see Scoring).

Dataset assembly reduced the stored rows to one row per trial, where a trial is the
combination of model, item, condition, and repetition. A trial had more than one row when
it had been retried. In that case a valid row was preferred over an invalid one, and among
valid rows the one with the lowest attempt number was kept. Among the analyzed models,
\valNattemptTwo{} of the kept rows come from a second attempt and \valNattemptThree{} from a
third, and the rest from the first. In the full sample of \valNComplete{} models,
\valNTruncSample{} trials, all from the reason condition of \valNTruncModels{} model, were
still cut off at the output limit after the last attempt. That model was among those the
awareness scan excluded, so every trial of the \valNAnalyzed{} analyzed models yielded a
valid answer.

\subsection{Scoring}

Each stored reply was first split into a reasoning trace and an answer text, and the
numeric answer was read from the answer text alone. When the server returned the
reasoning in a separate field, that field was the trace and the message content was the
answer text. Otherwise the reply was split at the last end-of-reasoning marker it
contained, such as \texttt{</think>}, with everything before the marker treated as the
trace and everything after it as the answer text. A reply with no reasoning markers at
all, which is the normal case in the intuition condition, was treated as answer text in
full. A reply that opened a reasoning block but never closed it had no answer text and
yielded no answer. Within the answer text, the answer was the value of the requested
JSON object when the whole text was that object, and otherwise the value of the last
\texttt{"answer"} field it contained. When the answer text contained no such field, the
answer was the last standalone number in it. Parsed answers were rounded to the nearest
integer before matching. The answer is read only after the reasoning
block because reasoning traces can restate the problem's numbers and try out candidate
answers, including the lure, before settling on one. A rule that searched the whole reply
for the first answer-like string would score such a candidate as the answer. The last
answer is taken because a model that writes a provisional answer and then revises it has
committed to the revised one.

Two scores were computed from the parsed numeric answers, following the two response
frequencies that \citet{meyer-etal-2024} report for every item. \emph{Accuracy} is the
fraction of trials whose answer exactly matches the item's correct value.
\emph{Intuitive-lure rate} is the fraction of trials whose answer exactly matches the
item's published MCE, the dominant intuitive lure. The two measures are not
complementary: an answer can be wrong without being the lure, so the lure rate
specifically indexes capture by the engineered intuitive response rather than error in
general. Each model's score in a condition aggregates \valNTrialsPerCell{} trials
(\valNItems{} items $\times$ \valNReps{} repetitions). Inference across models is
described under Statistical Analysis.

\subsection{Awareness Scan}

Because CRT items have correct answers, performance could in principle reflect
memorization rather than problem solving: the items were published in 2024 and may be
partly present in training corpora \citep{deng-etal-2024}. We therefore screened every model for textual
evidence that it \emph{recognizes} the items as CRT or studied stimuli, as opposed to
simply solving them. The scan reads the models' own reasoning text. In the reason
condition a model normally emits its chain of thought before the final answer, and
recognition, when present, surfaces there. The scan read the full stored reply, reasoning
field included, of every attempt of every CRT trial from all \valNComplete{} models in the
sample, invalid attempts included, and matched it against case-insensitive patterns.

The central design decision is where to draw the line between recognition and a
structural reaction to the item itself. Each CRT item is engineered around a dominant
intuitive lure that tempts the respondent into a simple but inappropriate operation
\citep{meyer-etal-2024}, so a careful solver \emph{should} pause, suspect a trick, and
notice the tempting-but-wrong answer. Accordingly, suspecting a trick, calling an item
a ``classic riddle,'' or predicting the tempting answer were classified as structural
reactions and never counted as awareness. What cannot be derived from the item's
structure is knowledge of the item as a \emph{studied stimulus}. Five signals were
counted. (a)~The model names the test (``cognitive reflection test,'' ``CRT''). (b)~It
connects an item to one of the three original CRT items \citep{frederick-2005}
(bat-and-ball, machines-and-widgets, or lily pads), which identifies the CRT lineage.
(c)~It names the authors, the 2024 study, or its foundational literature. (d)~It
attributes the item's published MCE value \citep{meyer-etal-2024} to people (``most people
answer 30''). This signal requires a population claim, so it is stricter than merely
predicting a tempting answer (``the intuitive answer is 30''), which is structural,
although the two are separated only by that population framing rather than by
demonstrated empirical knowledge. As a pattern match, signal D can also fire on hedged or
speculative statements about what people answer, so it errs toward exclusion. (e)~It explicitly claims a prior encounter with the item
or its source. A model was flagged as aware if any counted signal appeared on any of its CRT
trials, and all flagged models were excluded from every analysis. The scan is a
conservative lower bound, because it can only see awareness a model verbalizes and the
intuition condition normally emits little or no reasoning text. It is also a pattern-matching heuristic, so
verbatim matched snippets are archived for every flagged signal of every flagged model
(with the study data), and each flagged model can be audited by hand. Full signal definitions, the excluded structural signals, and
the matched evidence appear in the Supplementary Materials.

\subsection{Statistical Analysis}

\textit{Dataset assembly.} Trials that were invalid or whose reasoning was truncated
before completion were not included, so each model's scores in each condition are computed
over its valid trials. A model entered the analysis if it was in the sample and was not flagged by the
awareness scan. All analyses reported below were run on this dataset, and the awareness
exclusions are reported at the head of the Results. Analyses were run in Python with NumPy
and SciPy (versions in the Open Practices Statement).

\textit{Across-model inference.} The unit of analysis is the model: each model
contributes one accuracy and one lure rate per condition (its mean over that
condition's trials), and the two conditions are paired within model. Condition means
across models are reported with 95\% bootstrap confidence intervals (percentile method,
\valNBoot{} resamples of models, fixed seed~0). The condition effect is the per-model
difference (reason minus intuition), summarized by its mean and bootstrap CI and tested
with two distribution-free paired tests: a two-sided exact sign test on the direction of
each model's difference and a two-sided Wilcoxon signed-rank test on the nonzero differences (reported
when at least six models differ). Because many models are size or version variants
within a family, these tests treat correlated units as independent and their $p$-values
are anticonservative. We therefore emphasize direction counts and confidence intervals
over $p$-values, and report an organization-collapsed direction test among the
robustness analyses. No formal multiple-comparisons correction was applied: the
inference rests on the directional main effects, whose $p$-values fall far below any
plausible corrected threshold, and the secondary correlations and subset analyses are
reported as descriptive. Scaling analyses correlate each model's paired difference with
$\log_{10}$ of its total parameter count (Pearson and Spearman), and the same
non-independence caveat applies to their $p$-values. The share of intuition-condition
errors that were the lure, and each item's most common wrong answer, are computed over the
pooled trial responses rather than as per-model means. The non-lure error rate is the
share of trials whose answer was neither correct nor the lure (one minus accuracy minus
the lure rate). A relative reduction divides the drop in a condition mean, from intuition
to reason, by its intuition value. Because each (model, item, condition) cell was sampled \valNReps{} times under stochastic
decoding, a model's answer can vary across repetitions. Repetition consistency measures
how stable that answer is: for each cell we take the share of the \valNReps{} repetitions
giving the most common (modal) answer, then average over the \valNItems{} items per (model,
condition). It indexes whether the condition differences reflect systematic changes in
what the model answers rather than sampling variability.

\textit{Robustness analyses.} Models at perfect accuracy (or at zero lure rate) in
both conditions contribute structural zeros to the paired effects, so every effect was
recomputed on the corresponding headroom subset (models below ceiling under
intuition for accuracy, and models with at least one intuition-condition lure response
for the lure rate), together with a normalized version of each effect: the fraction
of a model's available headroom closed by reasoning,
$(\text{reason} - \text{intuition})/(1 - \text{intuition})$, and the fraction of its
intuition-condition lures eliminated. As a check on family non-independence, the
paired direction tests were also repeated with models collapsed to one unit per
developing organization.

\textit{Human benchmarks.} The human comparison values were computed from the raw data
of \citet{meyer-etal-2024} (data from \url{https://osf.io/9j8mc}). Using the authors' own item key and scoring rules, we
scored the attention-check-passing sample ($N = \valHumanN$) on the same two measures as
the models (accuracy and intuitive-lure rate), separately for the first exposure and
the second chance, with standard errors across participants. Following the authors'
convention, items answered correctly on first exposure are carried forward as correct
(and non-lure) on the second chance, which was administered only for items initially
missed. The re-analysis reproduces the published results: computed first-exposure
accuracy was \valHumanAccFirstEight/8 against the published \valHumanAccFirstPub/8,
computed second-chance accuracy was \valHumanAccSecondEight/8 against the published
\valHumanAccSecondPub/8, and every per-item solution and MCE rate matched its published
value to within \valHumanMaxDev. The item-to-column mapping, answer keys, and
attention-check criterion are reproduced from the authors' analysis script and ship with
our released code, so the human values are reproducible from the public Open Science Framework data alone.
Because, by the carry-forward convention, items answered correctly on first exposure
enter the second-chance proportion as fixed correct (and non-lure), the second-chance
standard errors are conditioned on that design.

\newpage
\section{Results}

\subsection{Awareness Scan and Exclusions}

The awareness scan examined every stored attempt of the \valScanTrials{} CRT trials from the
\valScanModels{} models in the sample (\valScanCharsM{} million characters of text).
Of the \valScanModels{} models, \valScanAware{} (\valScanAwarePct\%) showed at least one
counted awareness signal. By signal, \valSigA{} models named the CRT or the cognitive
reflection test, \valSigB{} connected an item to one of the three original CRT items,
\valSigD{} attributed an item's published most common error to people,
\valSigE{} made an explicit prior-encounter claim, and \valSigC{} of \valScanModels{} named
the authors, the 2024 study, or its foundational literature. Recognition of the source study itself was
absent (signal C), and connections to a named test pointed at the classic CRT lineage
(for instance, via the drill-and-hammer item's resemblance to the original
bat-and-ball). Structural reactions that the items are engineered to provoke were more common and
were not counted. In all, \valStructTrick{} models voiced trick suspicion, \valStructTemplate{}
called an item a generic classic riddle, and \valStructLure{} predicted the tempting answer
without any population claim. The \valStructOnly{} models that showed only such structural
signals were retained.

The \valNAware{} flagged models were removed from the \valNComplete{} models in the sample,
leaving \valNAnalyzed{} models and \valNTrials{} valid trials in the analysis set. Every
analysis reported below excludes the awareness-flagged models.

\subsection{Effect of Reasoning on Accuracy}

\begin{figure}[htbp]
\caption{Reasoning Raises CRT Accuracy: LLM Reasoning Effect Beside the Human Dual-Response Effect}
\label{fig:crt-main-accuracy}
\centering
\includegraphics[width=0.85\textwidth]{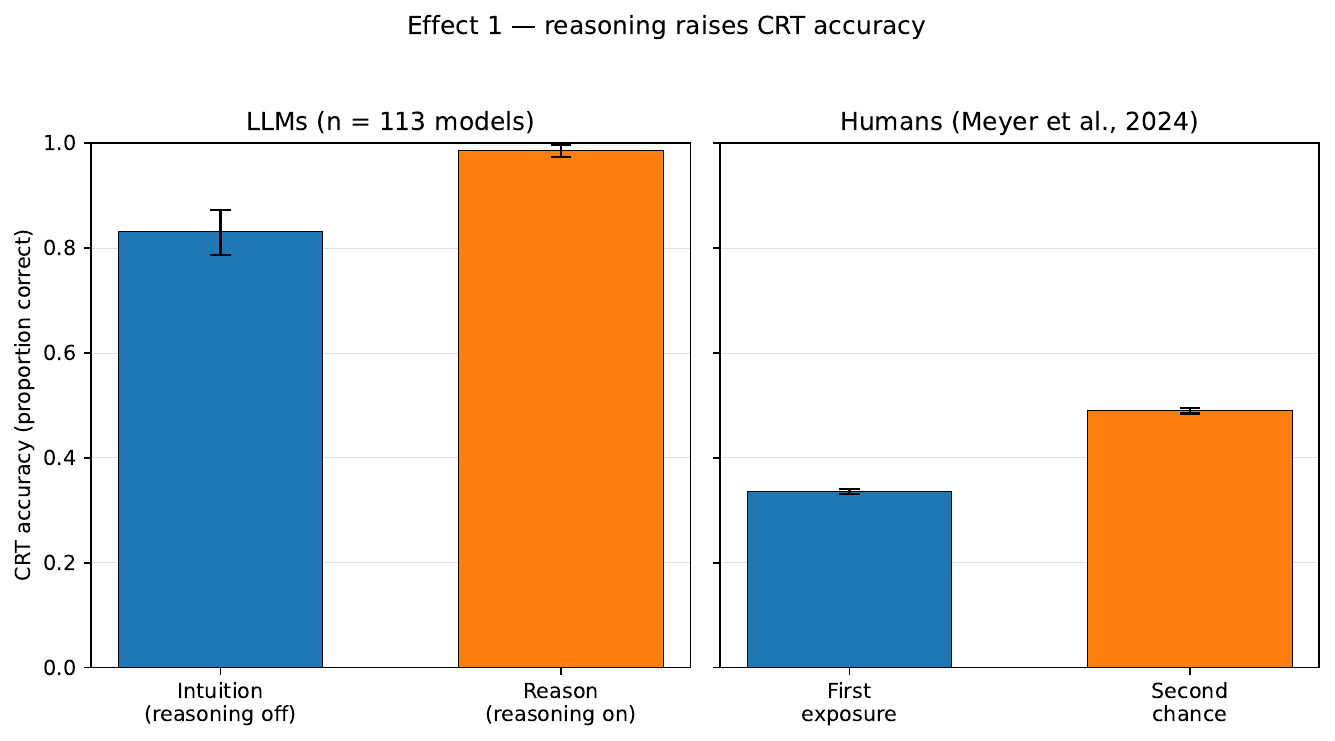}
\caption*{\textit{Note}. Left panel: mean CRT accuracy (proportion correct) across the
\valNAnalyzed{} analyzed models under the intuition (reasoning off) and reason (reasoning on)
conditions. Error bars are 95\% bootstrap confidence intervals across models. Right
panel: human accuracy on the same eight items at first exposure and on the second
chance, computed from the raw data of \citet{meyer-etal-2024} ($N = \valHumanN$). Error
bars are standard errors across participants. The two panels' error bars therefore
differ in kind and are not directly comparable in width. The same agent answers twice in
both panels. Intuition is the analogue of first exposure and reason of the second chance.}
\end{figure}

Figure~\ref{fig:crt-main-accuracy} shows the first main effect beside its human
counterpart. Across the \valNAnalyzed{} models, mean CRT accuracy was \valAccIntMean{},
95\% CI $[\valAccIntLo, \valAccIntHi]$, under intuition and \valAccReaMean{}, 95\% CI
$[\valAccReaLo, \valAccReaHi]$, under reason. The paired, within-model effect of
enabling reasoning was a mean gain of $\valAccDeltaMean$ (median $\valAccDeltaMedian$),
95\% CI $[\valAccDeltaLo, \valAccDeltaHi]$. In all, \valAccNPos{} models improved, none
declined, and \valAccTie{} were unchanged (exact sign test, $p \valAccSignP$, and Wilcoxon
signed-rank, $p \valAccWilcoxonP$). All \valAccTie{} unchanged models were already at
perfect accuracy in both conditions, so they contribute structural zeros to the mean. On
the non-independence of models as units, see Statistical Analysis and the
organization-collapsed test under Robustness. In the human benchmark, computed from the
raw data of \citet{meyer-etal-2024}, the same agents' accuracy rose from
\valHumanAccFirst{} ($= \valHumanAccFirstEight/8$, $SE = \valHumanAccFirstSE$) on first
exposure to \valHumanAccSecond{} ($= \valHumanAccSecondEight/8$, $SE =
\valHumanAccSecondSE$) on the second chance, a gain of $\valHumanAccDelta$.

The model and human effects share one structure: the same agent answers the same items
twice, first as an initial response and then after renewed deliberation, and accuracy
rises. For the models
the agent is literally unchanged between answers (identical weights, so identical
stored knowledge), and the toggle switches on deliberation together with the serial
test-time computation it entails (see Boundaries of the Inference). The two effects are
not parallel in level. The models' intuition baseline (\valAccIntMean) far exceeds the
human first-exposure baseline (\valHumanAccFirst), so the comparison is an analogy of
structure, not of level. At the same time, the intuition baseline sits below ceiling,
leaving about a sixth of the scale in headroom, so the direction of the reasoning gain is
cleanly interpretable. Its magnitude is a conservative, right-censored summary. The mean
averages over the \valNCeilingBoth{} models that could not register an effect at all, and
the reason-condition mean (\valAccReaMean) is itself pressed against the top of the scale.
The gain of $\valAccDeltaMean$ is therefore best read as a lower bound on what
deliberation contributes. The ceiling models are taken up in the scale analysis and in
Robustness to Ceiling and Floor Effects below.

\subsection{Effect of Reasoning on the Intuitive-Lure Rate}

\begin{figure}[htbp]
\caption{Reasoning Lowers the Intuitive-Lure Rate: LLM Reasoning Effect Beside the Human Dual-Response Effect}
\label{fig:crt-main-lure}
\centering
\includegraphics[width=0.85\textwidth]{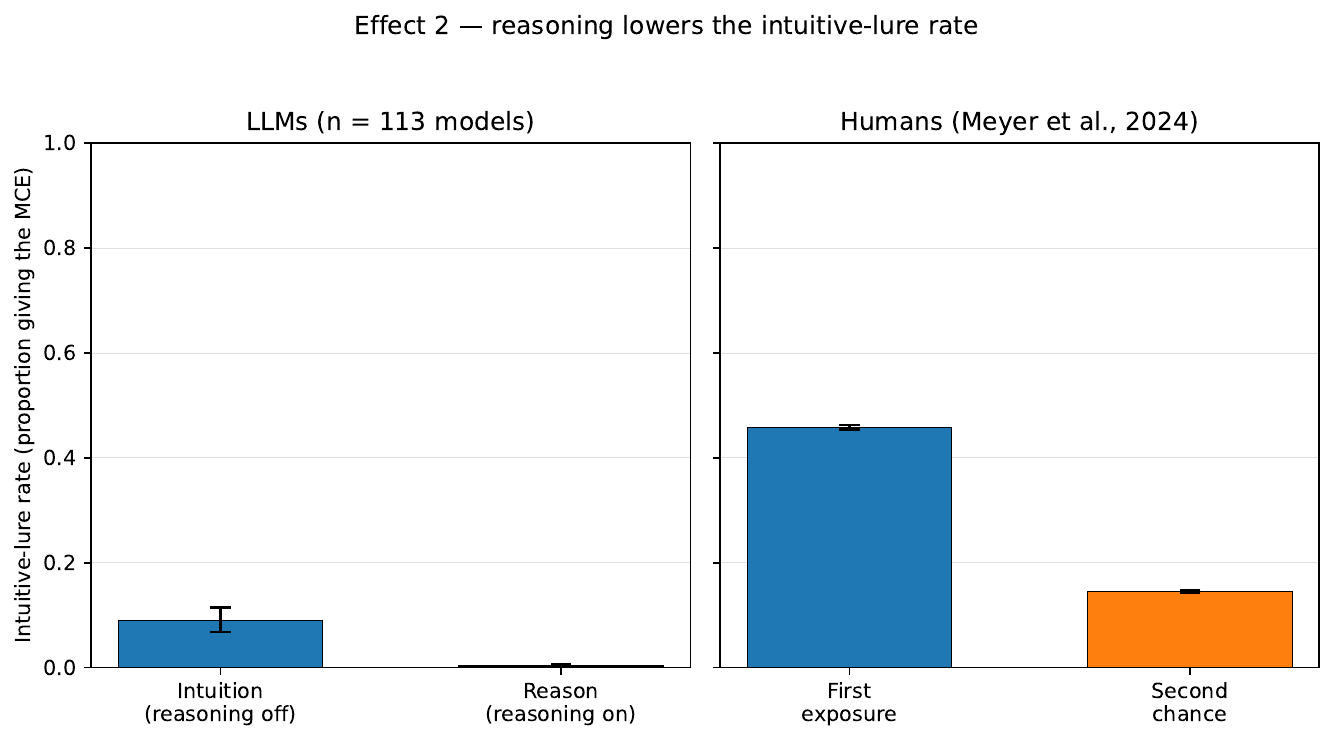}
\caption*{\textit{Note}. Left panel: mean intuitive-lure rate (proportion of answers
matching the published most common error) across the \valNAnalyzed{} analyzed models under the
intuition and reason conditions. Error bars are 95\% bootstrap confidence intervals
across models. Right panel: human lure rates on the same eight items at first exposure
and on the second chance, computed from the raw data of \citet{meyer-etal-2024}
($N = \valHumanN$). Error bars are standard errors across participants. The two panels'
error bars differ in kind and are not directly comparable in width.}
\end{figure}

The CRT's defining feature is not error in general but capture by a specific,
engineered intuitive answer, so the second main effect tracks the published most
common error directly (Figure~\ref{fig:crt-main-lure}). Under intuition, the models'
mean lure rate was \valLureIntMean{}, 95\% CI $[\valLureIntLo, \valLureIntHi]$. When
deliberation was off, the lure accounted for \valLureShareErrors{} of all wrong answers,
and on \valLureModalItems{} of the \valNItems{} items it was the most common wrong answer.
The error that most often captured the models was thus the answer that captures human
respondents (Table~\ref{tab:crt-items}). Under reason, the lure rate fell to \valLureReaMean{}, 95\% CI
$[\valLureReaLo, \valLureReaHi]$, a paired mean change of $\valLureDeltaMean$ (median
$\valLureDeltaMedian$), 95\% CI $[\valLureDeltaLo, \valLureDeltaHi]$. In all, \valLureNNeg{}
models decreased, none increased, and \valLureTie{} were unchanged (exact sign test,
$p \valLureSignP$, and Wilcoxon signed-rank, $p \valLureWilcoxonP$). All \valLureTie{}
unchanged models produced no lure response in either condition, which is why the median
change is zero. Errors that were not the lure declined as well, with the mean non-lure
error rate falling from \valNonLureIntMean{} to \valNonLureReaMean{}. The lure fell somewhat
more steeply in relative terms, however: the mean lure rate fell by \valRelLure{} of its
intuition-condition value, against \valRelNonLure{} for the mean non-lure error rate. As the per-item
analysis shows, the steepest lure drops landed on the items that captured models most
under intuition, which also have relatively high published human lure rates. The human benchmark shows the same signature. Lure
rates computed from the raw data of \citet{meyer-etal-2024} fell from \valHumanLureFirst{}
($SE = \valHumanLureFirstSE$) at first exposure to \valHumanLureSecond{} ($SE =
\valHumanLureSecondSE$) on the second chance, a change of $\valHumanLureDelta$.

As with accuracy, the parallel is structural rather than level-for-level. The models'
intuition-condition lure rate (\valLureIntMean) is only about a fifth of the human
first-exposure rate (\valHumanLureFirst). The near-floor baseline also bounds the
absolute change, because the models had at most \valLureIntMean{} of lure rate to lose,
where the human sample had \valHumanLureFirst{}. The models' smaller absolute drop
($\valLureDeltaMean$ vs.\ $\valHumanLureDelta$) is therefore consistent with the bound rather than
with weaker suppression. As a proportion of the baseline mean, the models shed \valRelLure{} of their lure
responses and the human respondents \valHumanRelLure{}. The effect among models with any lures to
lose is reported under Robustness to Ceiling and Floor Effects below. On direction the
two panels match: in both, deliberation reduces capture by the dominant engineered
lure, the mechanism the test was built around.

\subsection{Effect Size and Model Scale}

\begin{figure}[htbp]
\caption{Reasoning Effect as a Function of Model Size}
\label{fig:crt-scale}
\centering
\includegraphics[width=0.85\textwidth]{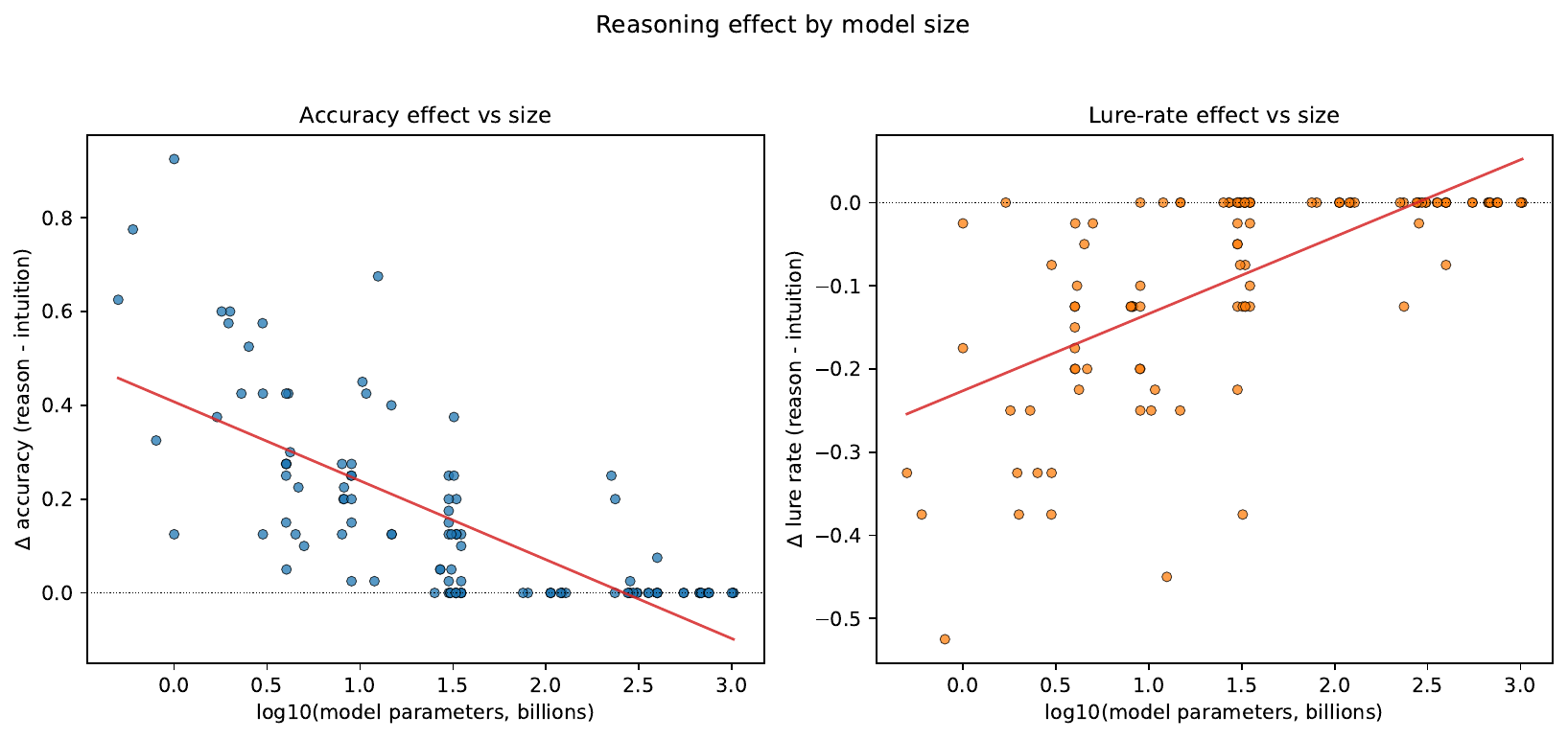}
\caption*{\textit{Note}. Each point is one model ($n = \valScaleN$). Left: per-model accuracy
effect (reason minus intuition) against $\log_{10}$ of total parameters (billions).
Right: per-model lure-rate effect on the same axis. Red lines are least-squares fits.
The dotted line marks a null effect. The points pinned to zero at large scale are
models at perfect accuracy (and zero lure rate) in both conditions, whose measured
effect is compressed against the ceiling. Each per-model effect is estimated from
\valNTrialsPerCell{} trials per cell and is treated as fixed by the least-squares fit
(sampling noise in the points is not modeled, but the consistency analysis indicates it is small).}
\end{figure}

All \valNAnalyzed{} analyzed models have a known parameter count, permitting a scaling analysis of
the reasoning effect (Figure~\ref{fig:crt-scale}). The per-model accuracy effect (reason
minus intuition) decreased reliably with model size: against $\log_{10}$ of total
parameters, Pearson $r = \valScaleAccR$, $p \valScaleAccP$, and Spearman $\rho = \valScaleAccRho$, $p \valScaleAccRhoP$
($n = \valScaleN$). The lure-rate effect showed the complementary pattern. Its magnitude
shrank with size, the signed effect becoming less negative (Pearson $r = \valScaleLureR$,
$p \valScaleLureP$, and Spearman $\rho = \valScaleLureRho$, $p \valScaleLureRhoP$). This
correlation carries the \valNZeroLureBoth{} models with no lure response in either
condition as exact zeros (see Robustness below).

The shrinking absolute effect must be read against the ceiling. The intuition-condition
baseline itself rises with scale ($r = \valBaseAccR$), steadily consuming the room in which a
gain could register. Of the \valNAnalyzed{} models, \valNCeilingBoth{}, concentrated among the
largest, were at perfect accuracy with zero lure responses in both conditions, so their
measured effect is necessarily zero whatever deliberation contributes. In
Figure~\ref{fig:crt-scale} these models appear as the cluster of points pinned to the
zero line at the right edge of each panel. Ceiling compression of this structural kind
is not the whole account, because the absolute gain declines with size even among models
below ceiling ($r = \valAccRoomScaleR$). An absolute trend on a bounded measure cannot, by
itself, say whether deliberation contributes less at scale, however. With reason-condition
accuracy near constant at ceiling, the accuracy effect is approximately the negative of
the intuition baseline, so its decline with size largely follows from that baseline
rising with size. The robustness analysis below addresses the question directly by
normalizing each model's gain by its remaining headroom. On that measure the trend
reverses, with larger models closing a \emph{greater} share of what remains.

\subsection{Response Consistency Across Repetitions}

\begin{figure}[htbp]
\caption{Distribution of Response Consistency Across the \valNReps{} Repetitions}
\label{fig:crt-consistency}
\centering
\includegraphics[width=0.85\textwidth]{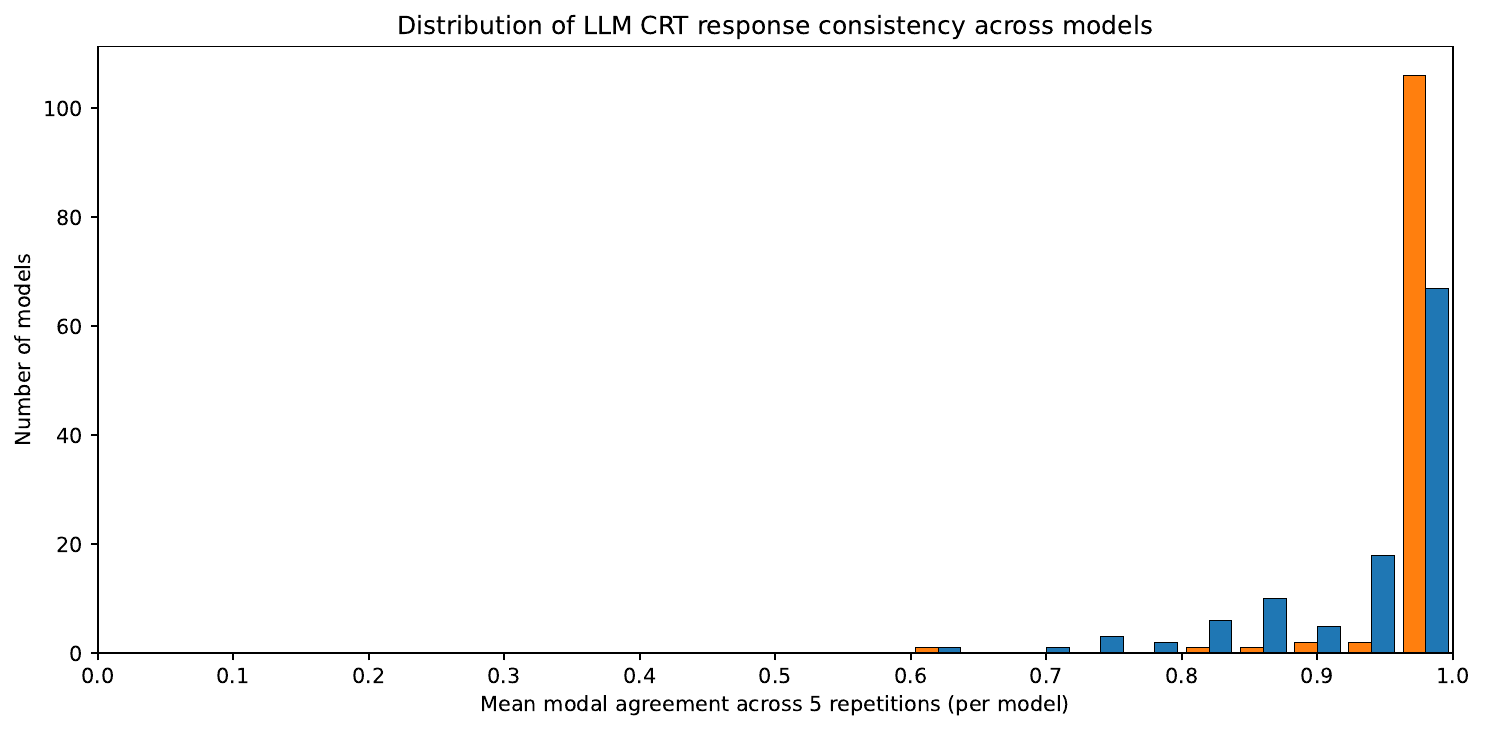}
\caption*{\textit{Note}. Histogram of per-model mean modal agreement (the modal
answer's share of a cell's five repetitions, averaged over the eight items) under
reason (orange) and intuition (blue), with $n = \valNAnalyzed$ models per condition. Both
distributions concentrate near 1.0, with the reason condition more tightly so.}
\end{figure}

Because every (model, condition, item) cell was answered \valNReps{} times under sampling,
the repetitions index how stable each model's answer is. Consistency was high in both
conditions and higher under reason (Figure~\ref{fig:crt-consistency}). Mean modal
agreement (the modal answer's share of a cell's \valNReps{} repetitions, averaged over the
\valNItems{} items) was $M = \valConsIntMean$ ($SD = \valConsIntSD$) under intuition and $M = \valConsReaMean$
($SD = \valConsReaSD$) under reason, a shift of $\valConsShift$ ($n = \valNAnalyzed$ models per condition). The
share of models with perfect agreement on every item rose from \valConsPerfectInt{} under
intuition to \valConsPerfectRea{} under reason. Two points follow. First, the condition
effects reported above are not noise from unstable answering. Models give largely the same
answer on every repetition, so the intuition--reason differences indicate systematic
changes in what is answered, not response scatter. Second, the direction of the shift is
itself informative. The CRT items have a single correct answer, so the higher agreement
under reason is consistent with repeated deliberation converging on that answer (as the
accuracy and lure results above indicate). The somewhat looser agreement under intuition
fits single-pass responses distributed over the correct answer, the lure, and other
errors.

\subsection{Per-Item Effects}

\begin{figure}[htbp]
\caption{Per-Item Accuracy Under Intuition and Reason}
\label{fig:crt-per-item-accuracy}
\centering
\includegraphics[width=0.85\textwidth]{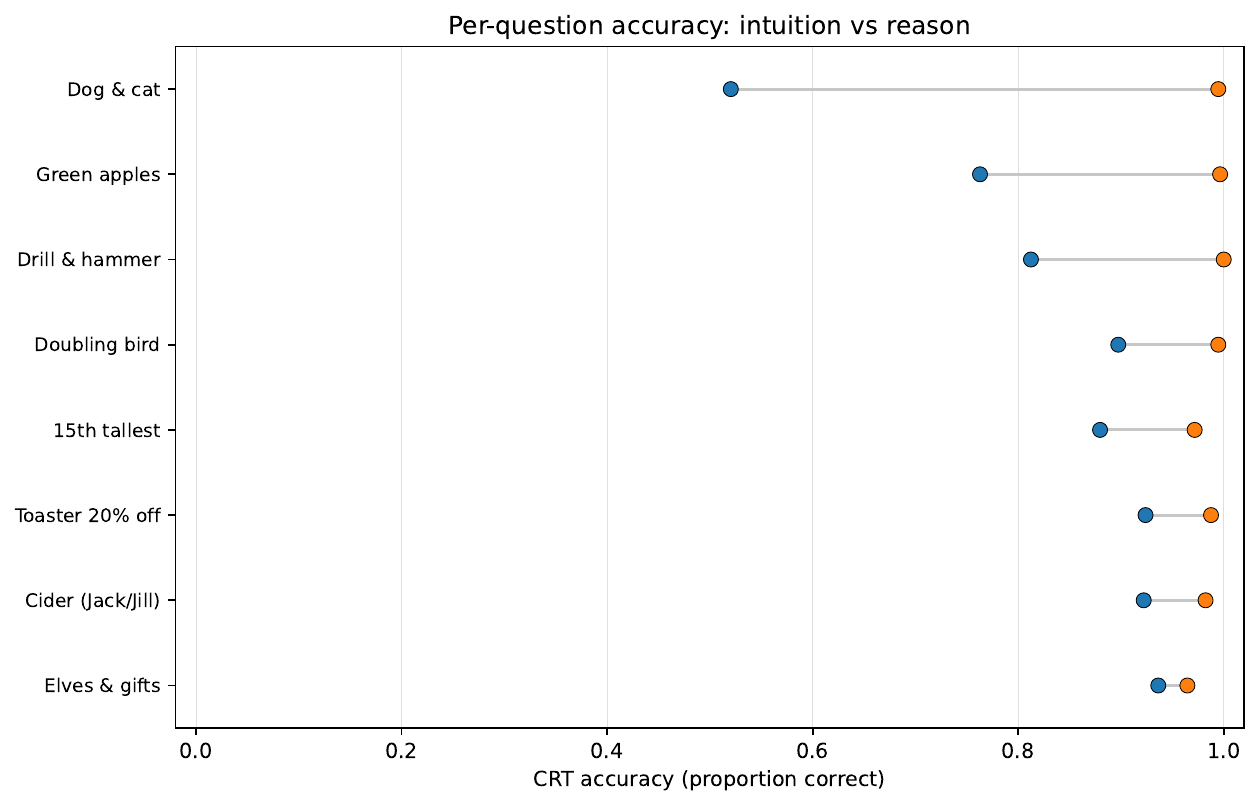}
\caption*{\textit{Note}. Each row is one CRT item. Blue marks the intuition condition
and orange the reason condition, averaged across the \valNAnalyzed{} models, with the connector
showing the within-item shift. Items are sorted by the size of the accuracy gain.}
\end{figure}

\begin{figure}[htbp]
\caption{Per-Item Intuitive-Lure Rate Under Intuition and Reason}
\label{fig:crt-per-item-lure}
\centering
\includegraphics[width=0.85\textwidth]{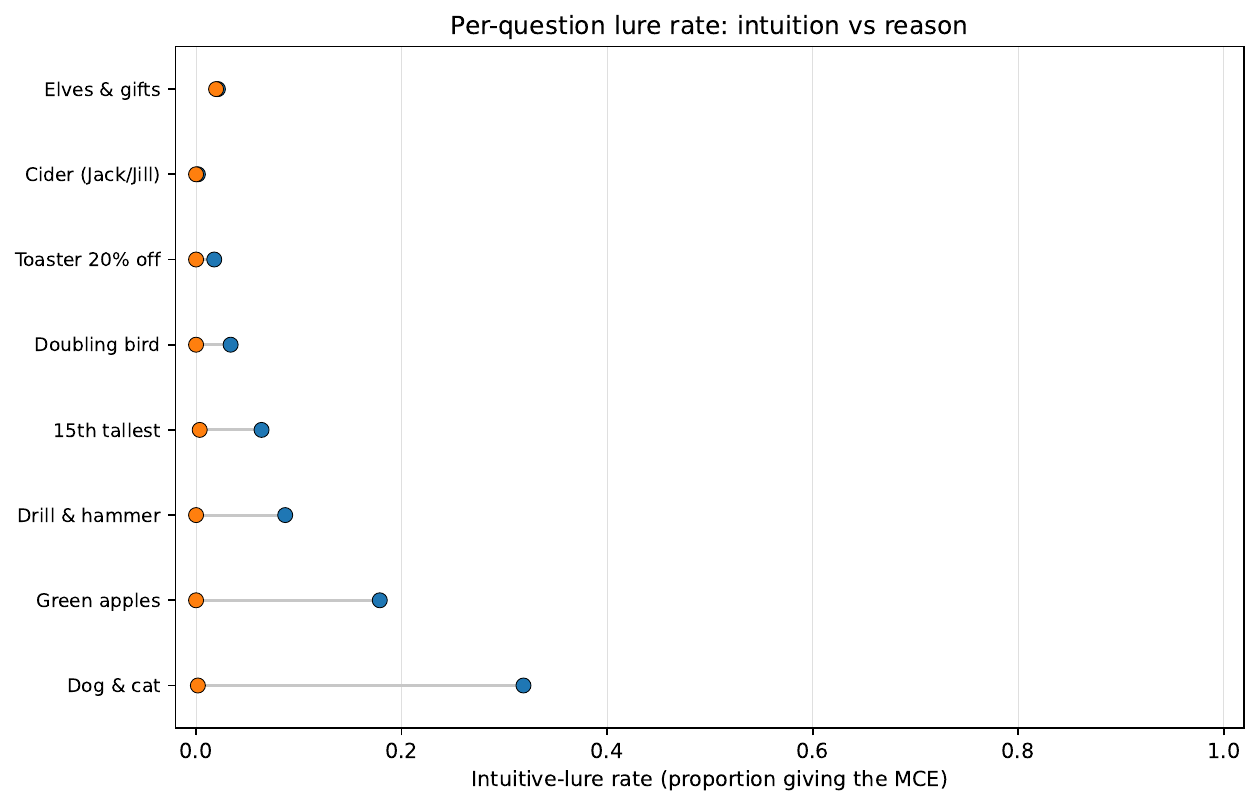}
\caption*{\textit{Note}. Same marker and connector convention as
Figure~\ref{fig:crt-per-item-accuracy}, for the intuitive-lure rate (proportion of
answers matching the published most common error), with items ordered by the change in
lure rate, the largest drop at the bottom.
The lure rate falls on every item, most steeply on the items that captured models
most under intuition (dog and cat, green apples).}
\end{figure}

Table~\ref{tab:crt-per-item} and Figures~\ref{fig:crt-per-item-accuracy} and
\ref{fig:crt-per-item-lure} break the two effects down by item. Accuracy rose on all
eight items, with gains ranging from $\valItemSixAccDelta$ (elves and gifts) to $\valItemTwoAccDelta$ (dog and
cat). The items that captured models most strongly under intuition were also the items
the toggle moved most. The dog-and-cat item drew the lure in \valItemTwoLureInt{} of models'
intuition responses (accuracy \valItemTwoAccInt), and reasoning cut its lure rate to
\valItemTwoLureRea{} (accuracy \valItemTwoAccRea). The green-apples item fell from a lure rate
of \valItemEightLureInt{} to \valItemEightLureRea{} (accuracy \valItemEightAccInt{} to
\valItemEightAccRea). The lure rate fell on every item, although on the items where it was
near floor under intuition the drop was small, as on the elves-and-gifts item
($\valItemSixLureDelta$, from \valItemSixLureInt{} to \valItemSixLureRea). The items with
the largest intuition-condition lure rates (dog and cat, green apples, and drill and
hammer) also have relatively high published human MCE rates
(\valItemTwoMCE{}, \valItemEightMCE{}, and \valItemOneMCE{}, respectively,
Table~\ref{tab:crt-items}). The correspondence is partial, however. The elves-and-gifts
item also has a high published MCE rate (\valItemSixMCE) but rarely drew the lure from the
models. Where models erred without deliberation, they tended to err on items that also
trap people, and the lure was their most common wrong answer on \valLureModalItems{} of
the \valNItems{} items. One reading caution applies to
the ordering of the accuracy gains. Reason-condition accuracy was confined to a narrow
near-ceiling band (\valItemAccReaMin{} to \valItemAccReaMax{} across the eight items), so
each item's gain is largely determined by its intuition-condition deficit. The ordering
therefore restates which items captured models under intuition rather than revealing
differential sensitivity to deliberation.

\begin{table}[htbp]
\caption{Per-Item Accuracy and Intuitive-Lure Rate by Condition, Averaged Across the \valNAnalyzed{} Models}
\label{tab:crt-per-item}
\centering
\begin{tabular}{@{}l rrr rrr@{}}
\toprule
 & \multicolumn{3}{c}{Accuracy} & \multicolumn{3}{c}{Lure rate} \\
\cmidrule(lr){2-4} \cmidrule(l){5-7}
Item & Intuition & Reason & $\Delta$ & Intuition & Reason & $\Delta$ \\
\midrule
Elves \& gifts & \valItemSixAccInt & \valItemSixAccRea & $\valItemSixAccDelta$ & \valItemSixLureInt & \valItemSixLureRea & $\valItemSixLureDelta$ \\
Cider (Jack/Jill) & \valItemSevenAccInt & \valItemSevenAccRea & $\valItemSevenAccDelta$ & \valItemSevenLureInt & \valItemSevenLureRea & $\valItemSevenLureDelta$ \\
Toaster 20\% off & \valItemFourAccInt & \valItemFourAccRea & $\valItemFourAccDelta$ & \valItemFourLureInt & \valItemFourLureRea & $\valItemFourLureDelta$ \\
15th tallest & \valItemFiveAccInt & \valItemFiveAccRea & $\valItemFiveAccDelta$ & \valItemFiveLureInt & \valItemFiveLureRea & $\valItemFiveLureDelta$ \\
Doubling bird & \valItemThreeAccInt & \valItemThreeAccRea & $\valItemThreeAccDelta$ & \valItemThreeLureInt & \valItemThreeLureRea & $\valItemThreeLureDelta$ \\
Drill \& hammer & \valItemOneAccInt & \valItemOneAccRea & $\valItemOneAccDelta$ & \valItemOneLureInt & \valItemOneLureRea & $\valItemOneLureDelta$ \\
Green apples & \valItemEightAccInt & \valItemEightAccRea & $\valItemEightAccDelta$ & \valItemEightLureInt & \valItemEightLureRea & $\valItemEightLureDelta$ \\
Dog \& cat & \valItemTwoAccInt & \valItemTwoAccRea & $\valItemTwoAccDelta$ & \valItemTwoLureInt & \valItemTwoLureRea & $\valItemTwoLureDelta$ \\
\bottomrule

\end{tabular}
\caption*{\textit{Note}. Cell values are means across the \valNAnalyzed{} analyzed models.
$\Delta$ = reason minus intuition. Items are sorted by the accuracy effect. Item
labels abbreviate the full item texts in Table~\ref{tab:crt-items}.}
\end{table}

\subsection{Per-Model Paired Effects}

\begin{figure}[htbp]
\caption{Per-Model Paired Comparison of the Two Conditions}
\label{fig:crt-paired}
\centering
\includegraphics[width=0.85\textwidth]{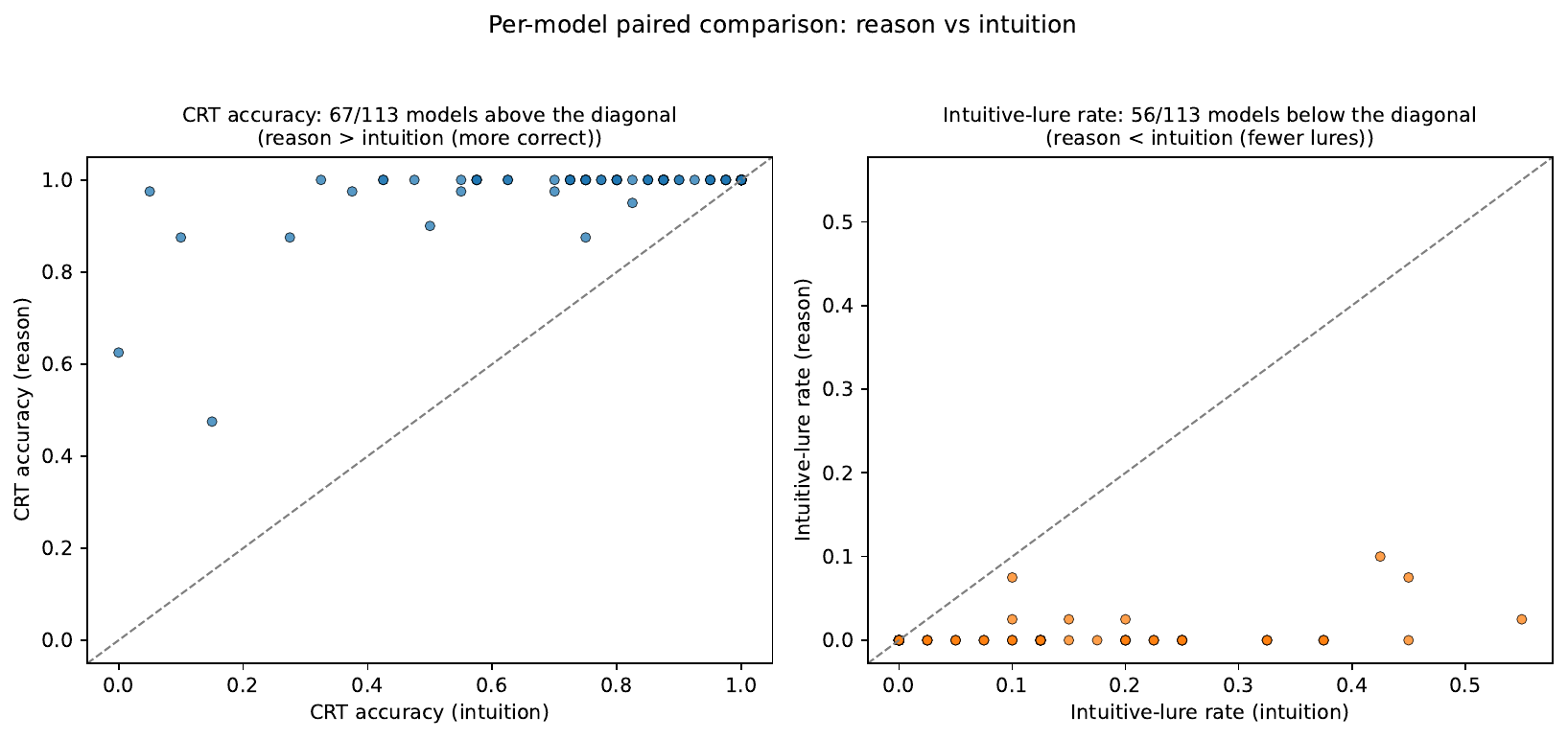}
\caption*{\textit{Note}. Each point is one model ($n = \valNAnalyzed$), with $x$ = intuition value,
$y$ = reason value, and the dashed diagonal marking no condition effect. Left: accuracy,
where points above the diagonal indicate higher accuracy under reason (\valAccAbove/\valNAnalyzed). Right:
intuitive-lure rate, where points below the diagonal indicate fewer lure responses under
reason (\valLureBelow/\valNAnalyzed). The dense cluster at the top-right of the accuracy panel comprises
the \valAccTie{} models at perfect accuracy in both conditions, and the stack at the origin of
the lure panel the \valLureTie{} models with no lure responses in either condition. Most of the
other points lie along the lure panel's floor, where reasoning eliminated the model's
intuition-condition lures.}
\end{figure}

Figure~\ref{fig:crt-paired} shows the paired comparison at the level of individual
models, with each model plotted against the identity line. For accuracy, \valAccAbove{} of \valNAnalyzed{}
models sat above the diagonal (more accurate under reason). The other \valAccTie{} showed no
change at all, because they were at perfect accuracy in both conditions and left the
toggle nothing to improve, and no model was less accurate under reason. For the lure
rate, \valLureBelow{} of \valNAnalyzed{} models sat below the diagonal (fewer intuitive lures
under reason), none sat above, and \valLureTie{} were unchanged. All \valLureTie{} unchanged
models produced no lure responses at all in either condition. The pattern is thus uniform
once headroom is accounted for. Enabling reasoning raised accuracy in every one of the
\valNBelowCeiling{} models not already at ceiling, and it reduced the lure rate in every one
of the \valNLureModels{} models that produced any lure responses under intuition. The
analyzed population spans model families from \valNOrgs{} organizations and several
template arguments for the toggle. The next subsection quantifies both the headroom
pattern and the organization-collapsed effect.

\subsection{Robustness to Ceiling and Floor Effects}

The mean effects reported above mix two kinds of models: those with room to move and
those without. Of the \valNAnalyzed{} models, \valNCeilingBoth{} were at perfect accuracy in both
conditions, and all \valNCeilingBoth{} also produced zero lure responses in both. No model that
was perfect under intuition lost that perfection under reason. On the lure side, \valNZeroLureInt{}
models produced no lure response under intuition at all, leaving deliberation nothing
to suppress. These structurally fixed models enter the paired means as exact zeros,
so the headline estimates understate the effect of deliberation wherever deliberation
had anything to do.

Restricting the analyses to models with measurable headroom makes this explicit.
Among the \valNBelowCeiling{} models below ceiling under intuition, mean accuracy rose from \valAccRoomIntMean{}, 95\%
CI $[\valAccRoomIntLo, \valAccRoomIntHi]$, to \valAccRoomReaMean{}, 95\% CI $[\valAccRoomReaLo, \valAccRoomReaHi]$, a paired gain of $\valAccRoomDeltaMean$
(median $\valAccRoomDeltaMedian$), 95\% CI $[\valAccRoomDeltaLo, \valAccRoomDeltaHi]$. Expressed as a fraction of each model's
available headroom, deliberation closed on average \valClosedMean{} of what remained, 95\% CI
$[\valClosedLo, \valClosedHi]$. The median below-ceiling model closed \emph{all} of its remaining
headroom, and \valClosedFull{} of the \valNBelowCeiling{} reached perfect accuracy under reason. Among the \valNLureModels{}
models that produced any lure responses under intuition, the mean lure rate fell from
\valLureRoomIntMean{}, 95\% CI $[\valLureRoomIntLo, \valLureRoomIntHi]$, to \valLureRoomReaMean{}, 95\% CI $[\valLureRoomReaLo, \valLureRoomReaHi]$, a change of $\valLureRoomDeltaMean$,
95\% CI $[\valLureRoomDeltaLo, \valLureRoomDeltaHi]$. All \valNLureModels{} decreased
(exact sign test, $p \valLureRoomSignP$). On average deliberation eliminated \valElimMean{} of a model's
intuition-condition lures, 95\% CI $[\valElimLo, \valElimHi]$, and \valElimFull{} of the \valNLureModels{} models eliminated
every one. One caution attaches to these restricted estimates. Membership in a headroom
subset is defined from the same observed scores being compared, so regression to the
mean could contribute to them. That contribution is bounded by sampling noise over the
\valNTrialsPerCell{} trials per cell, which is small given the high repetition consistency
reported above, and it cannot produce the near-total headroom closure observed.

The same normalization reframes the scale trend of Figure~\ref{fig:crt-scale}. The
absolute accuracy effect shrinks with model size even among below-ceiling models
(Pearson $r = \valAccRoomScaleR$, $p \valAccRoomScaleP$, $n = \valNBelowCeiling$), as it must
when the intuition baseline, and with it the available headroom, rises with scale
($r = \valBaseAccR$, $p \valBaseAccP$, $n = \valNAnalyzed$). The scale trend in the lure effect weakens but
persists once the \valNZeroLureInt{} no-lure models are set aside
($r = \valLureRoomScaleR$, $p \valLureRoomScaleP$, $n = \valNLureModels$). Measured against
what remains available, however, the effect \emph{grows} with scale. The fraction of
headroom closed correlates positively with $\log_{10}$~parameters (Spearman
$\rho = \valClosedScaleRho$, $p \valClosedScaleRhoP$, and Pearson $r = \valClosedScaleR$,
$p \valClosedScaleP$), as does the fraction of intuition lures eliminated
($\rho = \valElimScaleRho$, $p \valElimScaleP$, and Pearson $r = \valElimScaleR$,
$p \valElimScalePearP$). Larger models gain less in absolute terms because less remains to
gain, while converting a larger share of what does remain.

Finally, because many analyzed models are size or version siblings within a family,
we collapsed the \valNAnalyzed{} models to one unit per developing organization (\valOrgN{} organizations)
and re-ran the direction tests on the
per-organization mean effects. The accuracy effect was positive in \valOrgAccPos{} of the \valOrgN{}
organizations, exactly zero in the other \valOrgAccZero{}, and negative in none (exact sign
test, $p \valOrgAccSignP$, and Wilcoxon $p \valOrgAccWilcoxonP$). The lure effect was
negative in \valOrgLureNeg{} organizations, exactly zero in the other \valOrgLureZero{}, and
positive in none (exact sign test,
$p \valOrgLureSignP$, and Wilcoxon $p \valOrgLureWilcoxonP$). The conclusions therefore do not depend
on treating related models as independent units.

\newpage
\section{Discussion}

\subsection{Deliberation at Fixed Weights}

The Introduction closed with two directional hypotheses, and the data answered both.
Enabling reasoning raised mean CRT accuracy from \valAccIntMean{} to \valAccReaMean{} (a paired, within-model
gain of $\valAccDeltaMean$), and it lowered the intuitive-lure rate, the fraction
of answers matching each item's published most common error, from \valLureIntMean{} to \valLureReaMean{}
($\valLureDeltaMean$). Both movements run in the same direction as the human dual-response
benchmark computed from the raw data of \citet{meyer-etal-2024}, in which the same
respondents improved from \valHumanAccFirst{} to \valHumanAccSecond{} in accuracy and dropped from \valHumanLureFirst{} to \valHumanLureSecond{} in
lure rate between first exposure and second chance. Across a population of \valNAnalyzed{} models,
the same agent answering the same items twice (deliberation off, then on)
reproduced, in direction, the reflection signature that the test was built to elicit.

Two conclusions follow from the design, before any appeal to statistical reconstruction. The first
concerns dissociability. In the human work, separating reflection from mathematical
ability required statistical reconstruction (structural equation models over
lure-bearing and matched non-lure items) because people who differ in reflection
differ in numeracy, motivation, and much else besides \citep{meyer-etal-2024}. Here the
separation is constructive rather than inferential: the model's weights are
byte-for-byte identical in the two conditions, the item text is byte-identical, and the
sampling parameters are pinned, so nothing about what the network has stored (its
weights, hence its knowledge) distinguishes the cells. What differs
is whether the model performs serial, test-time deliberation (generating
intermediate tokens it then re-reads) before committing to an answer. We state the contrast precisely as \emph{deliberation versus
none at fixed weights}, not as ``ability held constant while disposition varies'':
for a language model, serial test-time computation is itself a constituent of task
ability, a point we return to under Boundaries of the Inference. What the design shows
is that engaging that computation, with the underlying network unchanged, is what
converts lure captures into correct answers wherever the items leave room to register
the change. This is within-agent support, strongest in the suppression of the lure, for the
claim that the CRT taps something separable from baseline mathematical competence.

The second conclusion concerns the purity of the signal. A human CRT score is an
unresolvable composite: numeracy, the disposition to check, and prior exposure to the
items all load on the same number
\citep{toplak-etal-2011, sinayev-peters-2015, stieger-reips-2016}, as does the
respondent's motivation on the day, since the deliberate operations the test rewards
require effort and motivation \citep{frederick-2005}.
The within-model toggle removes the between-person confounds wholesale (the
``respondents'' being compared are the same network), which no between-person human design can
arrange. One honest qualifier: enabling reasoning also adds computation, so the contrast
isolates the \emph{engagement of deliberation}, not a disposition separated from the
processing it requires. The signal is purer than a
human CRT score. It is not a measurement of reflective disposition in the
trait-psychology sense.

\subsection{From Latent Trait to Manipulable Cause}

In the human literature, cognitive reflection is a between-person latent trait. It is
measured, not assigned: no experimenter can instruct one group of participants to
\emph{be} reflective and another not to be, so the construct is recovered indirectly:
from individual differences \citep{frederick-2005}, from structural models that
separate a reflection factor from mathematical ability \citep{meyer-etal-2024}, and from
the dual-response procedure in which the same person answers, is told which items were
missed, and then reconsiders them \citep{meyer-etal-2024}. However ingenious, these designs remain correlational at the
level of the trait itself: reflective people differ from impulsive people in many ways
besides reflection.

The reasoning toggle converts that latent trait into an experimental factor realized
within a single agent. With reasoning disabled, the model is its own non-reflective
respondent: it reports whatever its single forward pass produces. With reasoning
enabled, the same agent (the same weights, reading the same item text) behaves as
a reflective respondent, deliberating before it answers. The two-cell experiment that is
impossible in humans is thereby run \valNAnalyzed{} times, once per model, and its result is not a
correlation but a difference: engaging deliberation, together with the serial
computation it entails, \emph{causes} the CRT improvement wherever the items leave
room to register one, with the agent's knowledge and weights fixed by construction. Among the models not already at perfect
accuracy in both conditions, the manipulation moved accuracy upward in \valAccRoomNPos{} of \valNBelowCeiling{} cases
(restricted-set estimates, and the selection caveat that attaches to them, appear under
Robustness to Ceiling and Floor Effects). The inference that the test's signature
reflects deliberation, rather than something that merely travels with deliberative
people, does not have to be extracted from a covariance structure here. It is read
off the manipulation directly.

\subsection{Lure Suppression as the Diagnostic Signature}

The CRT's defining engineering feature is not difficulty but the dominant intuitive
lure: each item is built so that a specific wrong answer arrives first and feels right
\citep{frederick-2005, meyer-frederick-2023}. A deliberation manipulation could, in
principle, improve scores without engaging that feature at all. A system that simply
``computed better'' when given more steps would shed errors of every kind alike. The
data show three signatures beyond that. First, under intuition the published most common error was the most common wrong answer on
\valLureModalItems{} of the \valNItems{} items and made up \valLureShareErrors{} of all wrong
answers. The dog-and-cat item drew the lure on \valItemTwoLureInt{}
of intuition responses and the green-apples item on \valItemEightLureInt{}, and the items that captured
models most strongly (dog and cat, green apples, drill and hammer) also have relatively
high published human MCE rates (\valItemTwoMCE{}, \valItemEightMCE{}, and \valItemOneMCE{}, respectively, see Table~\ref{tab:crt-items}),
although the item-level correspondence is partial (see Per-Item Effects). Second, although errors
of both kinds declined when reasoning was enabled, the lure did not simply ride a
general error decline: its relative suppression (\valRelLure{} of the intuition-condition rate
eliminated) was somewhat larger than that of non-lure errors (\valRelNonLure{}). Third, the lure rate fell on every
item, collapsing to \valLureReaMean{} overall, with the steepest drops on precisely
the items that had captured models most (although, with the reason-condition rate near floor
on every item, each item's drop largely mirrors its intuition-condition rate).

This is the part of the result that does the theoretical work. \citet{meyer-etal-2024}
argue that what the CRT measures is the disposition to override a tempting initial
response, and \citet{meyer-frederick-2023} locate the mechanism in the formation and
subsequent revision of intuitions, while finding that, in people, the erroneous intuition
often survives induced reflection. On this account, lure responses are informative in
their own right, marking capture by the engineered intuition rather than mere
inability, at least in able samples. The toggle reproduces the signature of that mechanism, not just its
outcome: deliberation (together with the serial computation it entails) acted somewhat more
strongly on the engineered intuitive answer than on other errors (a descriptive contrast
of relative reductions, not a tested difference), and in the models the lure rarely
survived deliberation (a reason-condition lure rate of \valLureReaMean). Had we tracked
accuracy alone, the replication would have been consistent with a generic
more-compute-better-math reading. The paired movement of the lure rate (its
concentration under intuition and its somewhat steeper relative suppression under reason) is
what makes the effect resemble cognitive reflection in Meyer et al.'s sense.

\subsection{Spontaneous Reflection Without Error Feedback}

The replication also bears on a confound inside the human procedure it parallels. In
the dual-response paradigm of \citet{meyer-etal-2024}, the second chance is not a bare
invitation to think again: participants are told which items they answered incorrectly
before reconsidering. The human second-chance gain therefore bundles two ingredients:
re-engaged deliberation and an external error signal that marks where deliberation
should be aimed. From the human data alone, one cannot say how much of the improvement
is reflection and how much is feedback.

The reason condition supplies no feedback of any kind. The model is never told it erred,
never shown which items to revisit, and never given a second look at its first answer.
It is merely permitted to generate intermediate reasoning before responding. That the
gain reproduces in direction under this stricter manipulation (accuracy up, lure
rate down, across the model population) points to the engagement of \emph{spontaneous}
reflection as an active ingredient: deliberation that corrects the lure on
its own, without being pointed at the error (and, as under Boundaries of the Inference,
inseparable from the serial computation it adds). The two designs are thus complementary arms of one
argument. The human dual response shows that reflection-plus-feedback moves answers off
the lure, whereas the toggle shows that deliberation without feedback does. Their convergence makes
re-engaged deliberation, rather than the error feedback, the most plausible shared ingredient. This also sharpens the parallel with human
two-response procedures in which deliberation is engaged without external correction
\citep{bago-deneys-2017, thompson-etal-2011}: the model's reasoning pass functions as an
un-cued second response (engaged without an external error signal), and its effect on
accuracy runs in the same direction as the net shift toward correct answers that
\citet{bago-deneys-2017} observed between initial and final responses.

\subsection{Robustness, Ceiling Compression, and the Dose--Response Question}

The effect is not the artifact of one model, one vendor, or one engineering choice. The
\valNAnalyzed{} analyzed models come from \valNOrgs{} organizations, span more than three orders of magnitude
in scale, and include dense and mixture-of-experts architectures. Most importantly for
interpretation, they realize the reasoning toggle through several distinct chat-template
arguments, from boolean thinking switches to graded effort levels. Training recipes,
alignment pipelines, and reasoning-mode implementations differ across these families in
many details, yet no model showed a reversal of the paired effect. The replications are not
independent (the models share a transformer substrate, heavily overlapping training
corpora, and in some cases literal base-model ancestry), but beyond that shared
inheritance, what the analyzed implementations have in common is the abstract contrast
the design isolates: answering from a single pass versus answering after serial
self-generated context. The survival of the effect across this partially correlated
variation is therefore most consistent with the shared deliberative contrast (serial
self-conditioning together with the computation it entails) as the operative
factor rather than anything idiosyncratic to one family, and the effect does not depend
on counting related models as independent
evidence: collapsed to one unit per organization, the accuracy effect remained positive
in \valOrgAccPos{} of \valOrgN{} organizations and the lure effect negative in \valOrgLureNeg{} with none positive.

The only moderator we analyzed, model scale, appears to mark the instrument's boundary
rather than deliberation's. The absolute accuracy effect shrank reliably with parameter
count ($r = \valScaleAccR$), and still did among models below ceiling ($r = \valAccRoomScaleR$), but
that is how a bounded measure must behave when the baseline itself rises with scale
($r = \valBaseAccR$), consuming the room in which any gain could register. Of the \valNAnalyzed{}
models, \valNCeilingBoth{}, concentrated among the largest, had no room at all, sitting at perfect accuracy
with zero lures in both conditions. The normalized analyses favor reading
the trend as the instrument's boundary. Measured as the share of remaining headroom closed, the effect
\emph{rose} with scale ($\rho = \valClosedScaleRho$), as did the share of intuition lures
eliminated ($\rho = \valElimScaleRho$). In every model the items could still measure,
enabling deliberation helped (\valAccRoomNPos{} of \valNBelowCeiling{} models for accuracy, and \valLureRoomNNeg{} of \valNLureModels{} with intuition
lures), and larger models converted more of what remained. On these items, then, the
shrinking absolute effect is most parsimoniously read as ceiling compression rather than
a real attenuation of deliberation with capability, because the normalized effect does not
decline, and if anything rises, with scale. Probing deliberation at the
frontier will require a harder successor instrument, much as it did for humans when
the classic items grew too familiar \citep{haigh-2016, stieger-reips-2016}.

The heterogeneity of toggle implementations also exposes a question the human paradigm
cannot ask. Meyer et al.'s procedure \citep{meyer-etal-2024} is binary (one first
exposure, one second chance), and human deliberation cannot be dosed as cleanly. Model deliberation can: some of the
analyzed models expose graded effort levels (used here only as on or off settings), and
performance depends on test-time computation \citep{agarwal-etal-2025} and on the reasoning
budget \citep{ghosal-etal-2025}, though not always monotonically. Whether lure suppression grows smoothly
with the amount of deliberation (a dose--response curve of lure rate against
reasoning tokens) or arrives as a step once any deliberation is engaged is, to our
knowledge, open, and the present design extends naturally to it.

\subsection{The Lure Rate as Built-In Evidence Against Recall}

For language models trained on web-scale text, the contamination worry is the analogue
of the ``burned'' bat-and-ball: as the classic items saturated the culture, a correct
answer stopped guaranteeing reflection in humans \citep{haigh-2016, stieger-reips-2016},
and benchmark leakage can inflate model scores \citep{deng-etal-2024}, in ways that
survive even paraphrase of the leaked items \citep{yang-etal-2023}. The design addressed this risk
prospectively (a recent instrument, an awareness scan with exclusions), but the results
also address it internally, because the two response measures react to memorization in
opposite ways.

A model answering from a memorized key would have no reason to fall for the lure when
deliberation is off: retrieval of a stored answer predicts errorless recall in both
conditions, and it cannot explain why withholding deliberation reinstates the
human most common errors, mostly on items that also have relatively high human MCE
rates. The elevated lure rate under intuition is therefore positive evidence that the
models producing it (\valNLureModels{} of the \valNAnalyzed{}) were genuinely
(mis)processing the items, producing the same simple-but-inappropriate
response that people give, rather than reciting their solutions. For the models that never gave the lure, this
check is silent. The intuition lure rate thus serves as internal evidence against a
pure-recall account, without an external contamination analysis.

The awareness scan complements this from the other side, subject to its stated limit
that it sees only verbalized recognition. Recognition was uncommon (\valScanAware{} of \valScanModels{} screened
models, all excluded), never named the source study, and item-level recognition, where
it occurred, was mostly classic-lineage references or assertions of the documented human
error (see Results). Recognition of this kind, rather than of the source study, is consistent with the profile expected if the updated stimuli are scarce in training
corpora while the classic CRT is abundant \citep{meyer-etal-2018}. Together, the lure-capture pattern and the scan make
the replication more consistent with reflection than with recall. Neither, though, can
exclude contamination that leaves no verbal trace (silent memorization of an item's
answer, or exposure to paraphrased variants \citep{yang-etal-2023}) because such recall need not
reinstate the lure under intuition. The decisive test would administer
fresh, unpublished items that preserve each item's lure-generating structure but change
its surface and its correct answer, so that any memorized answer mapping is useless.
Whether the reasoning effect replicates on such variants is left to future work.

\subsection{Dual-Process Organization Beyond the Human Substrate}

Dual-process theory is, in humans, an inference from behavior: the fast--slow
architecture is postulated to explain response patterns, conflict signals, and
two-response dissociations, not observed directly \citep{evans-stanovich-2013,
pennycook-etal-2015b, bago-deneys-2017, thompson-etal-2011}. In the present system the architecture is not postulated but
implemented, exposed, and switchable. One network either answers in a single
bounded pass or loops its own intermediate products back through itself before
answering, and an experimenter can flip between the two organizations at will. That
toggling this implementation reproduces the human reflection signature (accuracy
rising and the engineered lure suppressed somewhat more steeply than other errors, on items that also trap
human respondents) is converging evidence that the System-1/System-2 organization can
be realized in a \emph{class of computational systems}, those that pair fast parallel
pattern completion with optional serial computation over their own outputs, rather
than a quirk of human psychology.

This reading has independent support. \citet{moskovitz-etal-2024} derive a
dual-controller organization normatively, showing that a cheap default process and a
costly overriding one emerge from a single optimization pressure in paired neural
networks. \citet{hagendorff-etal-2023b} found that earlier large language models fall for
CRT-style intuitive lures, the larger ones even more often than people do (a tendency their most capable models
had learned to avoid) and that chain-of-thought reasoning helps models escape the
traps. The present results extend that picture with the within-model
manipulation: the same network exhibits both profiles, and which one appears is decided
by whether serial self-conditioning is engaged. We state the conclusion as functional
convergence, not mechanistic identity. The claim is that the input--output dissociation
characteristic of dual-process cognition (lure capture without deliberation, lure
override with it) does not require a human substrate. It is not that transformer
inference and human neurobiology implement deliberation the same way, and the claim is
about computational organization, not origin. The models' fast-pass intuitions are
themselves compiled from next-token prediction over human text.

\subsection{The Transparent Mechanism as a Hypothesis Generator}

So far the analogy has run from humans to models. Here we run it backwards. The model's
deliberative mechanism is specified and inspectable in a way the brain's is not, so it can be
taken apart and asked what reflection actually requires. Doing so suggests a number of
hypotheses about human deliberation, offered here as speculation for future work,
not as claims that the present study can settle.

The mechanism is worth restating because its transparency is the asset. The model's
``System 1'' is a single forward pass that emits the answer directly. Its ``System 2''
is the \emph{same network} generating intermediate tokens that are written into its own
context and re-ingested, step by step, before the answer. There is no second module, no
separate faculty: the only difference between the conditions is whether the network
feeds its own intermediate outputs back to itself. Treating this system as a
transparent model organism for deliberation suggests the following hypotheses about the
human case.

\begin{itemize}

\item \emph{System 2 as System 1 iterated.} One network plus a feedback loop sufficed
to reproduce the reflection dissociation: lure capture without deliberation, override
with it. By parsimony, human deliberation may likewise be the
intuitive machinery applied recursively to its own intermediate products rather than a
second system. Inner speech, which has been proposed as a medium of thought
\citep{kompa-2024}, is one candidate for such a product.

\item \emph{Deliberation as serial time traded for depth.} A single pass has limited
depth and, when it errs, tends to fall to the nearest attractor, the lure. Each generated token adds
another cheap step of serial depth, a trade that is provably real, in expressive power, for transformers
\citep{merrill-sabharwal-2024}. In humans: intuition is shallow parallel
pattern-matching, reflection the serial working-memory steps that push past its limit
\citep{evans-stanovich-2013}.

\item \emph{Override, not prevention.} When the fast pass errs on a CRT item, the lure
is its characteristic output. It is what the same network produces, on those items,
when the loop is off. Deliberation plausibly does not stop
the intuition from forming but constructs a derivation whose conclusion outranks it.
This matches what human respondents show: the lure exerts measurable pull even on those
who answer correctly \citep{travers-etal-2016}, and it is the
content of the formation-and-revision account \citep{meyer-frederick-2023}.

\item \emph{Initiation as the bottleneck.} Once the loop runs, the models almost always
solve the items (reason-condition accuracy was \valAccReaMean{}), so what separates the
conditions is not the network's stored competence (its weights are unchanged) but whether the serial pass is engaged at all. Humans look
similar: much of the second-chance gain in \citet{meyer-etal-2024} comes from errors
corrected as soon as reconsideration is prompted (albeit with feedback marking the
errors, whereas the toggle result shows that, for models, the engagement without feedback suffices). ``Low reflection'' may be a failure to
spend serial computation, not the lack of a faculty, which would be why the CRT
dissociates from a comparably difficult math test \citep{meyer-etal-2024}, and why metacognitive
triggers of rethinking matter \citep{thompson-etal-2011}.

\item \emph{Language as the medium of serial thought.} The model's scratchpad is made
of tokens. That chaining steps \emph{in language} turns a fast pattern-matcher into a
reliable reasoner \citep{nye-etal-2021, wei-etal-2022} is consistent with
accounts of the phonological loop and of inner speech as a medium of thought
\citep{baddeley-1992, buchsbaum-2013, kompa-2024}.

\item \emph{A graded, testable transfer.} Because intermediate generation provably adds
expressive power \citep{merrill-sabharwal-2024} (though whether lure
suppression itself scales smoothly with reasoning tokens is, as noted under Robustness,
untested here), the mechanism predicts that human deliberation is
graded in serial steps, and that external-scratchpad interventions (writing the
problem out, added time, or the instruction to ``think step by step'' that helps
models \citep{kojima-etal-2022}) are predicted to produce CRT gains resembling the
toggle's, by adding depth rather than knowledge.

\end{itemize}

These hypotheses must be bounded. The model performs \emph{all} of its serial
computation externally, whereas humans surely command some internal serial capacity.
The LLM is the limiting case in which all serial work is externalized into re-read
tokens, which is informative about how deliberation is organized but says nothing about
the capacity of the human working-memory buffer. Its fast-pass intuitions are
compiled from next-token prediction over human text, while human intuitions arise from
evolution and experience, so the analogy holds at the level of computational
organization, not of origin or content. And humans may well recruit additional neural
and affective substrates in deliberation. The model shows such substrates are not
\emph{necessary} for the reflection dissociation (a sufficiency argument), not
that humans lack or do not use them.

\subsection{Boundaries of the Inference}

The first boundary is that predictive validity was not tested. A central claim in
\citet{meyer-etal-2024} is not merely that reflection is measurable but that CRT scores
predict beliefs, preferences, and judgments better than a comparably difficult test of
mathematical ability. This record of predictive and correlational findings is what the test's influence rests on
\citep{frederick-2005, toplak-etal-2011, pennycook-etal-2015}. The present study validates
the reflection \emph{manipulation}: it shows that engaging deliberation produces the
test's diagnostic signature. Whether a model's CRT score predicts anything else about
that model's judgments was not examined, and the bridge from our manipulation to
Meyer et al.'s predictive claims is interpretive, not measured.

The second boundary, the central one, concerns what ``held fixed'' means. The weights are fixed, but the
computation is not. The reason condition does not engage deliberation at matched
processing. It adds serial test-time computation, in the form of a longer output and
many more forward passes over the model's own intermediate tokens, and this is intrinsic
to the manipulation rather than a removable nuisance. The measured effect is therefore
\emph{deliberation-with-additional-computation versus none}, and the design cannot
distinguish the engagement of deliberation from the extra computation that deliberation
entails. Indeed, on the mechanistic view developed above the two may not be distinct
things: deliberation may simply \emph{be} the spending of serial computation. So the analogy is functional (the same agent doing more deliberative work), not
proof that a disposition, separate from the processing it drives, moves the score. We regard this as the main interpretive
limit of the study.

Third, the two populations are aligned in structure, not in level. The models'
intuition-condition accuracy (\valAccIntMean) is more than double the human first-exposure
accuracy (\valHumanAccFirst), and their intuition lure rate (\valLureIntMean) is roughly a fifth of the human
one (\valHumanLureFirst). The side-by-side panels in the Results are therefore a directional analogy
(same agent, deliberation off then on, same signature), not a matched contrast,
and no formal quantitative comparison of effect sizes across species of agent, beyond
the descriptive contrasts reported in the Results, is licensed.
Because the models' intuition baseline sits below ceiling, the \emph{direction} of
the reasoning effect is cleanly interpretable. Its magnitude, however, remains compressed. The
reason-condition mean (\valAccReaMean) is effectively at ceiling and the reason-condition lure rate
(\valLureReaMean) near floor, so the reported deltas are lower bounds, not calibrated effect
sizes. Relatedly, the human second chance includes error feedback while the reason
condition does not \citep[see above and][]{meyer-etal-2024}. The parallel is between
analogous, not identical, manipulations.

Two further limits concern inference and measurement. The analyzed models are not independent units: many
are size or version variants within a family (successive Qwen or GLM releases,
for example), so the across-model tests treat correlated observations as independent
and their $p$-values are anticonservative. The organization-collapsed check bounds the
damage: with the \valNAnalyzed{} models reduced to one unit per developing organization, both effects survive at
conventional levels (accuracy positive in \valOrgAccPos{} of \valOrgN{} organizations, lure negative in \valOrgLureNeg{}
with none positive), so the conclusions do not rest on the inflated unit count,
though residual dependence through shared training corpora and distillation lineages
cannot be excluded by any grouping. The second limit is the one this paper can least
escape: the instrument is bounded, and much of the model population sits near its
bounds. Specifically, \valNCeilingBoth{} models had no accuracy headroom at all, \valNZeroLureInt{} had no intuition lures
to lose, and already-easy items (the elves-and-gifts lure rate was near floor under
intuition) register little movement whatever deliberation does. The headroom-restricted
and normalized analyses indicate that this compression \emph{understates} rather than
manufactures the effect (the below-ceiling gain is larger than the headline mean,
and the median below-ceiling model closed all of its remaining headroom), but the
flip side is a real measurement limit: for the largest models, the eight items are too
easy to measure deliberation at all, and quantitative claims about the size of the
reasoning effect at the frontier are not licensed by these data. Any residual familiarity
with the 2024 items that escaped the awareness scan would plausibly act in the same direction,
compressing rather than inflating the reported effect.

\newpage
\section{Conclusion}


We administered the updated eight-item CRT of \citet{meyer-etal-2024} to a large,
heterogeneous population of open-weight language models, each answering every item
twice, once with reasoning disabled and once with it enabled, at fixed weights
and byte-identical item text. Enabling deliberation raised accuracy and
suppressed the intuitive lure (somewhat more steeply, in relative terms, than other errors),
most strongly where the bounded instrument left room to
register it, reproducing the direction of the human first-exposure-to-second-chance
effect \citep{meyer-etal-2024}. An awareness scan found recognition of the items to be uncommon, mostly references to the
classic CRT lineage or assertions of the documented human error, and never directed at the
source study.

Beyond supporting the claim that reflection is separable from baseline mathematical
competence \citep{meyer-etal-2024}, and furnishing a signal free of the between-person confounds that human
CRT scores carry, the replication does six things. It converts cognitive reflection
from a latent between-person trait into a manipulable, within-agent cause. It
reproduces the signature of the test's specific mechanism (suppression of the engineered
lure somewhat beyond a general reduction of error) rather than a generic accuracy gain. It
bears on the spontaneous, un-cued character of the reflection, because the reason
condition removes the error feedback embedded in the human second-chance procedure. It provides converging evidence that dual-process organization
can be realized outside the human substrate (in systems that pair fast pattern
completion with optional serial computation), a claim about organization, not origin,
since models trained on human text inherit human-shaped intuitions. It makes the
intuition lure rate internal evidence against pure recall. And because the mechanism is
exactly specified (one network looped through its own tokens), it runs the analogy
in reverse, yielding a parsimonious hypothesis for future work: that human System 2
may be System 1 iterated through a serial, language-scaffolded working-memory loop,
trading time for computational depth, and overriding, rather than erasing, the
intuitive attractor.

\newpage
\section{Declarations}

\subsection{Funding}
This research did not receive any specific grant from funding agencies in the public, commercial, or not-for-profit sectors.

\subsection{Acknowledgements}
This research was enabled in part by computing resources provided by the Digital Research Alliance of Canada.

\subsection{Competing Interests}
The author has no competing interests to declare.

\subsection{Ethics Approval and Consent}
Not applicable. No new human data were collected. The human benchmark is a secondary analysis of the publicly released, anonymized data of \citet{meyer-etal-2024}.

\subsection{Availability of Data and Materials}
All collected model responses, the CRT trial dataset, and the per-model analysis outputs are available on Zenodo (see the Open Practices Statement).

\subsection{Code Availability}
All code used to collect the responses, assemble the dataset, run the awareness scan and the analyses, and generate the figures and the reported values is available on Zenodo (see the Open Practices Statement).

\subsection{Use of Large Language Models}
During the preparation of this work the author used various Claude (Anthropic) and GPT (OpenAI) models to search and summarize the literature, write and review the code for data collection and analysis, draft and revise the text of the manuscript under the author's direction, and check its citations, statistics, and claims against the cited sources and the data. The study's design, hypotheses, analytic decisions, and interpretations are the author's. After using these tools, the author reviewed and edited the content as needed and takes full responsibility for the content of the published article.

\section{Open Practices Statement}

The collected model responses, the CRT trial dataset, the awareness-scan report with its matched evidence, the analysis code, the per-model outputs, and every figure in this article are available on Zenodo \citep{yanitski-2026-crt-data}. The repository also contains the per-model list of serving settings and the scripts that ran the models. The reanalyzed human data of \citet{meyer-etal-2024} are publicly available at \url{https://osf.io/9j8mc} and are not redistributed. Newly collected models were served with vLLM 0.30.0 or a September 2026 build of llama.cpp, and all analyses ran in Python 3.14 (NumPy 2.4, SciPy 1.17) with a fixed bootstrap seed. The study was not preregistered.

\newpage
\bibliography{references}

@article{frederick-2005,
  author  = {Frederick, S.},
  title   = {Cognitive reflection and decision making},
  journal = {Journal of Economic Perspectives},
  year    = {2005},
  volume  = {19},
  number  = {4},
  pages   = {25--42},
  note    = {\url{https://doi.org/10.1257/089533005775196732}}
}

@incollection{kahneman-frederick-2002,
  author    = {Kahneman, D. and Frederick, S.},
  title     = {Representativeness revisited: {A}ttribute substitution in intuitive judgment},
  booktitle = {Heuristics and biases: {T}he psychology of intuitive judgment},
  editor    = {Gilovich, T. and Griffin, D. and Kahneman, D.},
  publisher = {Cambridge University Press},
  year      = {2002},
  pages     = {49--81}
}

@incollection{kahneman-frederick-2005,
  author    = {Kahneman, D. and Frederick, S.},
  title     = {A model of heuristic judgment},
  booktitle = {The {Cambridge} handbook of thinking and reasoning},
  editor    = {Holyoak, K. J. and Morrison, R. G.},
  publisher = {Cambridge University Press},
  year      = {2005},
  pages     = {267--293}
}

@article{meyer-frederick-2023,
  author  = {Meyer, A. and Frederick, S.},
  title   = {The formation and revision of intuitions},
  journal = {Cognition},
  year    = {2023},
  volume  = {240},
  pages   = {105380},
  note    = {\url{https://doi.org/10.1016/j.cognition.2023.105380}}
}

@article{meyer-etal-2024,
  author  = {Meyer, A. and Attali, Y. and Bar-Hillel, M. and Frederick, S. and Kahneman, D.},
  title   = {Cognitive reflection is a distinct and measurable trait},
  journal = {Proceedings of the National Academy of Sciences},
  year    = {2024},
  volume  = {121},
  number  = {49},
  pages   = {e2409191121},
  note    = {\url{https://doi.org/10.1073/pnas.2409191121}}
}

@article{deneys-etal-2013,
  author  = {{De Neys}, W. and Rossi, S. and Houd{\'e}, O.},
  title   = {Bats, balls, and substitution sensitivity: {C}ognitive misers are no happy fools},
  journal = {Psychonomic Bulletin \& Review},
  year    = {2013},
  volume  = {20},
  number  = {2},
  pages   = {269--273},
  note    = {\url{https://doi.org/10.3758/s13423-013-0384-5}}
}

@article{travers-etal-2016,
  author  = {Travers, E. and Rolison, J. J. and Feeney, A.},
  title   = {The time course of conflict on the {Cognitive Reflection Test}},
  journal = {Cognition},
  year    = {2016},
  volume  = {150},
  pages   = {109--118},
  note    = {\url{https://doi.org/10.1016/j.cognition.2016.01.015}}
}

@article{szaszi-etal-2017,
  author  = {Szaszi, B. and Szollosi, A. and Palfi, B. and Aczel, B.},
  title   = {The cognitive reflection test revisited: {E}xploring the ways individuals solve the test},
  journal = {Thinking \& Reasoning},
  year    = {2017},
  volume  = {23},
  number  = {3},
  pages   = {207--234},
  note    = {\url{https://doi.org/10.1080/13546783.2017.1292954}}
}

@article{toplak-etal-2011,
  author  = {Toplak, M. E. and West, R. F. and Stanovich, K. E.},
  title   = {The {Cognitive Reflection Test} as a predictor of performance on heuristics-and-biases tasks},
  journal = {Memory \& Cognition},
  year    = {2011},
  volume  = {39},
  number  = {7},
  pages   = {1275--1289},
  note    = {\url{https://doi.org/10.3758/s13421-011-0104-1}}
}

@article{toplak-etal-2014,
  author  = {Toplak, M. E. and West, R. F. and Stanovich, K. E.},
  title   = {Assessing miserly information processing: {A}n expansion of the {Cognitive Reflection Test}},
  journal = {Thinking \& Reasoning},
  year    = {2014},
  volume  = {20},
  number  = {2},
  pages   = {147--168},
  note    = {\url{https://doi.org/10.1080/13546783.2013.844729}}
}

@article{pennycook-etal-2015,
  author  = {Pennycook, G. and Cheyne, J. A. and Barr, N. and Koehler, D. J. and Fugelsang, J. A.},
  title   = {On the reception and detection of pseudo-profound bullshit},
  journal = {Judgment and Decision Making},
  year    = {2015},
  volume  = {10},
  number  = {6},
  pages   = {549--563},
  note    = {\url{https://doi.org/10.1017/S1930297500006999}}
}

@article{pennycook-etal-2016,
  author  = {Pennycook, G. and Cheyne, J. A. and Koehler, D. J. and Fugelsang, J. A.},
  title   = {Is the cognitive reflection test a measure of both reflection and intuition?},
  journal = {Behavior Research Methods},
  year    = {2016},
  volume  = {48},
  number  = {1},
  pages   = {341--348},
  note    = {\url{https://doi.org/10.3758/s13428-015-0576-1}}
}

@article{baron-etal-2015,
  author  = {Baron, J. and Scott, S. and Fincher, K. and Metz, S. E.},
  title   = {Why does the {Cognitive Reflection Test} (sometimes) predict utilitarian moral judgment (and other things)?},
  journal = {Journal of Applied Research in Memory and Cognition},
  year    = {2015},
  volume  = {4},
  number  = {3},
  pages   = {265--284},
  note    = {\url{https://doi.org/10.1016/j.jarmac.2014.09.003}}
}

@article{liberali-etal-2012,
  author  = {Liberali, J. M. and Reyna, V. F. and Furlan, S. and Stein, L. M. and Pardo, S. T.},
  title   = {Individual differences in numeracy and cognitive reflection, with implications for biases and fallacies in probability judgment},
  journal = {Journal of Behavioral Decision Making},
  year    = {2012},
  volume  = {25},
  number  = {4},
  pages   = {361--381},
  note    = {\url{https://doi.org/10.1002/bdm.752}}
}

@article{campitelli-gerrans-2014,
  author  = {Campitelli, G. and Gerrans, P.},
  title   = {Does the cognitive reflection test measure cognitive reflection? {A} mathematical modeling approach},
  journal = {Memory \& Cognition},
  year    = {2014},
  volume  = {42},
  number  = {3},
  pages   = {434--447},
  note    = {\url{https://doi.org/10.3758/s13421-013-0367-9}}
}

@article{sinayev-peters-2015,
  author  = {Sinayev, A. and Peters, E.},
  title   = {Cognitive reflection vs. calculation in decision making},
  journal = {Frontiers in Psychology},
  year    = {2015},
  volume  = {6},
  pages   = {532},
  note    = {\url{https://doi.org/10.3389/fpsyg.2015.00532}}
}

@article{mastrogiorgio-2015,
  author  = {Mastrogiorgio, A.},
  title   = {Commentary: {C}ognitive reflection vs. calculation in decision making},
  journal = {Frontiers in Psychology},
  year    = {2015},
  volume  = {6},
  pages   = {936},
  note    = {\url{https://doi.org/10.3389/fpsyg.2015.00936}}
}

@inproceedings{welsh-etal-2013,
  author    = {Welsh, M. B. and Burns, N. R. and Delfabbro, P. H.},
  title     = {The {Cognitive Reflection Test}: {H}ow much more than numerical ability?},
  booktitle = {{Proceedings of the 35th Annual Meeting of the Cognitive Science Society}},
  publisher = {Cognitive Science Society},
  year      = {2013},
  pages     = {1587--1592}
}

@article{attali-bar-hillel-2020,
  author  = {Attali, Y. and Bar-Hillel, M.},
  title   = {The false allure of fast lures},
  journal = {Judgment and Decision Making},
  year    = {2020},
  volume  = {15},
  number  = {1},
  pages   = {93--111},
  note    = {\url{https://doi.org/10.1017/S1930297500006938}}
}

@article{otero-etal-2022,
  author  = {Otero, I. and Salgado, J. F. and Moscoso, S.},
  title   = {Cognitive reflection, cognitive intelligence, and cognitive abilities: {A} meta-analysis},
  journal = {Intelligence},
  year    = {2022},
  volume  = {90},
  pages   = {101614},
  note    = {\url{https://doi.org/10.1016/j.intell.2021.101614}}
}

@article{patel-etal-2019,
  author  = {Patel, N. and Baker, S. G. and Scherer, L. D.},
  title   = {Evaluating the cognitive reflection test as a measure of intuition/reflection, numeracy, and insight problem solving, and the implications for understanding real-world judgments and beliefs},
  journal = {Journal of Experimental Psychology: General},
  year    = {2019},
  volume  = {148},
  number  = {12},
  pages   = {2129--2153},
  note    = {\url{https://doi.org/10.1037/xge0000592}}
}

@article{levy-2023,
  author  = {Levy, N.},
  title   = {What does the {CRT} measure? {P}oor performance may arise from rational processes},
  journal = {Philosophical Psychology},
  year    = {2023},
  volume  = {36},
  number  = {1},
  pages   = {58--84},
  note    = {\url{https://doi.org/10.1080/09515089.2022.2038123}}
}

@article{sirota-etal-2021,
  author  = {Sirota, M. and Dewberry, C. and Juanchich, M. and Valu{\v{s}}, L. and Marshall, A. C.},
  title   = {Measuring cognitive reflection without maths: {D}evelopment and validation of the verbal cognitive reflection test},
  journal = {Journal of Behavioral Decision Making},
  year    = {2021},
  volume  = {34},
  number  = {3},
  pages   = {322--343},
  note    = {\url{https://doi.org/10.1002/bdm.2213}}
}

@article{meyer-etal-2018,
  author  = {Meyer, A. and Zhou, E. and Frederick, S.},
  title   = {The non-effects of repeated exposure to the {Cognitive Reflection Test}},
  journal = {Judgment and Decision Making},
  year    = {2018},
  volume  = {13},
  number  = {3},
  pages   = {246--259},
  note    = {\url{https://doi.org/10.1017/S1930297500007683}}
}

@article{haigh-2016,
  author  = {Haigh, M.},
  title   = {Has the standard {Cognitive Reflection Test} become a victim of its own success?},
  journal = {Advances in Cognitive Psychology},
  year    = {2016},
  volume  = {12},
  number  = {3},
  pages   = {145--149},
  note    = {\url{https://doi.org/10.5709/acp-0193-5}}
}

@article{stieger-reips-2016,
  author  = {Stieger, S. and Reips, U.-D.},
  title   = {A limitation of the {Cognitive Reflection Test}: {F}amiliarity},
  journal = {PeerJ},
  year    = {2016},
  volume  = {4},
  pages   = {e2395},
  note    = {\url{https://doi.org/10.7717/peerj.2395}}
}

@article{evans-stanovich-2013,
  author  = {Evans, J. St. B. T. and Stanovich, K. E.},
  title   = {Dual-process theories of higher cognition: {A}dvancing the debate},
  journal = {Perspectives on Psychological Science},
  year    = {2013},
  volume  = {8},
  number  = {3},
  pages   = {223--241},
  note    = {\url{https://doi.org/10.1177/1745691612460685}}
}

@article{bago-deneys-2017,
  author  = {Bago, B. and {De Neys}, W.},
  title   = {Fast logic?: {E}xamining the time course assumption of dual process theory},
  journal = {Cognition},
  year    = {2017},
  volume  = {158},
  pages   = {90--109},
  note    = {\url{https://doi.org/10.1016/j.cognition.2016.10.014}}
}

@article{thompson-etal-2011,
  author  = {Thompson, V. A. and {Prowse Turner}, J. A. and Pennycook, G.},
  title   = {Intuition, reason, and metacognition},
  journal = {Cognitive Psychology},
  year    = {2011},
  volume  = {63},
  number  = {3},
  pages   = {107--140},
  note    = {\url{https://doi.org/10.1016/j.cogpsych.2011.06.001}}
}

@article{pennycook-etal-2015b,
  author  = {Pennycook, G. and Fugelsang, J. A. and Koehler, D. J.},
  title   = {What makes us think? {A} three-stage dual-process model of analytic engagement},
  journal = {Cognitive Psychology},
  year    = {2015},
  volume  = {80},
  pages   = {34--72},
  note    = {\url{https://doi.org/10.1016/j.cogpsych.2015.05.001}}
}

@incollection{baddeley-hitch-1974,
  author    = {Baddeley, A. D. and Hitch, G.},
  title     = {Working memory},
  booktitle = {The psychology of learning and motivation},
  editor    = {Bower, G. H.},
  volume    = {8},
  pages     = {47--89},
  publisher = {Academic Press},
  address   = {New York, NY},
  year      = {1974}
}

@article{baddeley-1992,
  author  = {Baddeley, A. D.},
  title   = {Working memory},
  journal = {Science},
  year    = {1992},
  volume  = {255},
  number  = {5044},
  pages   = {556--559},
  note    = {\url{https://doi.org/10.1126/science.1736359}}
}

@article{buchsbaum-2013,
  author  = {Buchsbaum, B. R.},
  title   = {The role of consciousness in the phonological loop: {H}idden in plain sight},
  journal = {Frontiers in Psychology},
  year    = {2013},
  volume  = {4},
  pages   = {496},
  note    = {\url{https://doi.org/10.3389/fpsyg.2013.00496}}
}

@article{kompa-2024,
  author  = {Kompa, N. A.},
  title   = {Inner speech and `pure' thought --- {D}o we think in language?},
  journal = {Review of Philosophy and Psychology},
  year    = {2024},
  volume  = {15},
  number  = {3},
  pages   = {645--662},
  note    = {\url{https://doi.org/10.1007/s13164-023-00678-w}}
}

@article{moskovitz-etal-2024,
  author  = {Moskovitz, T. and Miller, K. J. and Sahani, M. and Botvinick, M. M.},
  title   = {Understanding dual process cognition via the minimum description length principle},
  journal = {PLOS Computational Biology},
  volume  = {20},
  number  = {10},
  pages   = {e1012383},
  year    = {2024},
  note    = {\url{https://doi.org/10.1371/journal.pcbi.1012383}}
}

@inproceedings{wei-etal-2022,
  author    = {Wei, J. and Wang, X. and Schuurmans, D. and Bosma, M. and Ichter, B. and Xia, F. and Chi, E. H. and Le, Q. V. and Zhou, D.},
  title     = {Chain-of-thought prompting elicits reasoning in large language models},
  booktitle = {{Advances in Neural Information Processing Systems (NeurIPS)}},
  volume    = {35},
  pages     = {24824--24837},
  year      = {2022}
}

@inproceedings{kojima-etal-2022,
  author    = {Kojima, T. and Gu, S. S. and Reid, M. and Matsuo, Y. and Iwasawa, Y.},
  title     = {Large language models are zero-shot reasoners},
  booktitle = {{Advances in Neural Information Processing Systems (NeurIPS)}},
  volume    = {35},
  year      = {2022}
}

@misc{nye-etal-2021,
  author       = {Nye, M. and Andreassen, A. J. and Gur-Ari, G. and Michalewski, H. and Austin, J. and Bieber, D. and Dohan, D. and Lewkowycz, A. and Bosma, M. and Luan, D. and Sutton, C. and Odena, A.},
  title        = {Show your work: {S}cratchpads for intermediate computation with language models},
  year         = {2021},
  howpublished = {arXiv preprint arXiv:2112.00114},
  note         = {\url{https://arxiv.org/abs/2112.00114}}
}

@inproceedings{merrill-sabharwal-2024,
  author    = {Merrill, W. and Sabharwal, A.},
  title     = {The expressive power of transformers with chain of thought},
  booktitle = {{International Conference on Learning Representations (ICLR)}},
  year      = {2024}
}

@inproceedings{aher-etal-2023,
  author    = {Aher, G. and Arriaga, R. I. and Kalai, A. T.},
  title     = {Using large language models to simulate multiple humans and replicate human subject studies},
  booktitle = {{Proceedings of the 40th International Conference on Machine Learning (ICML)}},
  series    = {Proceedings of Machine Learning Research},
  volume    = {202},
  year      = {2023}
}

@article{hagendorff-etal-2023b,
  author  = {Hagendorff, T. and Fabi, S. and Kosinski, M.},
  title   = {Human-like intuitive behavior and reasoning biases emerged in large language models but disappeared in {ChatGPT}},
  journal = {Nature Computational Science},
  year    = {2023},
  volume  = {3},
  number  = {10},
  pages   = {833--838},
  note    = {\url{https://doi.org/10.1038/s43588-023-00527-x}}
}

@misc{hagendorff-etal-2023a,
  author       = {Hagendorff, T. and Fabi, S. and Kosinski, M.},
  title        = {Thinking fast and slow in large language models},
  year         = {2023},
  howpublished = {Preprint},
  note         = {Preprint version of the article published in \emph{Nature Computational Science}}
}

@article{yax-etal-2024,
  author  = {Yax, N. and Anll\'{o}, H. and Palminteri, S.},
  title   = {Studying and improving reasoning in humans and machines},
  journal = {Communications Psychology},
  year    = {2024},
  volume  = {2},
  pages   = {51},
  note    = {\url{https://doi.org/10.1038/s44271-024-00091-8}}
}

@article{dillion-etal-2023,
  author  = {Dillion, D. and Tandon, N. and Gu, Y. and Gray, K.},
  title   = {Can {AI} language models replace human participants?},
  journal = {Trends in Cognitive Sciences},
  year    = {2023},
  volume  = {27},
  number  = {7},
  pages   = {597--600},
  note    = {\url{https://doi.org/10.1016/j.tics.2023.04.008}}
}

@article{macmillan-scott-musolesi-2024,
  author  = {{Macmillan-Scott}, O. and Musolesi, M.},
  title   = {({I}r)rationality and cognitive biases in large language models},
  journal = {Royal Society Open Science},
  year    = {2024},
  volume  = {11},
  number  = {6},
  pages   = {240255},
  note    = {\url{https://doi.org/10.1098/rsos.240255}}
}

@inproceedings{jones-steinhardt-2022,
  author    = {Jones, E. and Steinhardt, J.},
  title     = {Capturing failures of large language models via human cognitive biases},
  booktitle = {{Advances in Neural Information Processing Systems (NeurIPS)}},
  volume    = {35},
  year      = {2022}
}

@inproceedings{shaki-etal-2023,
  author    = {Shaki, J. and Kraus, S. and Wooldridge, M.},
  title     = {Cognitive effects in large language models},
  booktitle = {{ECAI} 2023: 26th {E}uropean {C}onference on {A}rtificial {I}ntelligence},
  editor    = {Gal, K. and Now{\'e}, A. and Nalepa, G. J. and Fairstein, R. and R{\u{a}}dulescu, R.},
  series    = {Frontiers in Artificial Intelligence and Applications},
  volume    = {372},
  pages     = {2105--2112},
  publisher = {IOS Press},
  year      = {2023},
  note      = {\url{https://doi.org/10.3233/FAIA230505}}
}

@inproceedings{ma-etal-2025,
  author    = {Ma, S. and Xie, W. and Wang, Z. and Sun, X. and Chen, K. and Wang, E. and Liu, W. and Tong, H.},
  title     = {Do large language models truly grasp mathematics? {A}n empirical exploration from cognitive psychology},
  booktitle = {{Proceedings of the 47th Annual Meeting of the Cognitive Science Society}},
  publisher = {Cognitive Science Society},
  year      = {2025},
  note      = {\url{https://escholarship.org/uc/item/24x9t7s1}}
}

@misc{agarwal-etal-2025,
  author       = {Agarwal, A. and Sengupta, A. and Chakraborty, T.},
  title        = {The art of scaling test-time compute for large language models},
  year         = {2025},
  howpublished = {arXiv preprint arXiv:2512.02008},
  note         = {\url{https://arxiv.org/abs/2512.02008}}
}

@inproceedings{ghosal-etal-2025,
  author    = {Ghosal, S. S. and Chakraborty, S. and Reddy, A. and Lu, Y. and Wang, M. and Manocha, D. and Huang, F. and Ghavamzadeh, M. and Bedi, A. S.},
  title     = {Does thinking more always help? {M}irage of test-time scaling in reasoning models},
  booktitle = {{Advances in Neural Information Processing Systems (NeurIPS)}},
  volume    = {38},
  year      = {2025},
  note      = {\url{https://doi.org/10.52202/085713-5747}}
}

@inproceedings{deng-etal-2024,
  author    = {Deng, C. and Zhao, Y. and Tang, X. and Gerstein, M. and Cohan, A.},
  title     = {Investigating data contamination in modern benchmarks for large language models},
  booktitle = {{Proceedings of the 2024 Conference of the North American Chapter of the Association for Computational Linguistics: Human Language Technologies (Volume 1: Long Papers)}},
  pages     = {8706--8719},
  year      = {2024}
}

@inproceedings{dong-etal-2024,
  author    = {Dong, Y. and Jiang, X. and Liu, H. and Jin, Z. and Gu, B. and Yang, M. and Li, G.},
  title     = {Generalization or memorization: {D}ata contamination and trustworthy evaluation for large language models},
  booktitle = {{Findings of the Association for Computational Linguistics: ACL 2024}},
  pages     = {12039--12050},
  year      = {2024}
}

@misc{yang-etal-2023,
  author       = {Yang, S. and Chiang, W.-L. and Zheng, L. and Gonzalez, J. E. and Stoica, I.},
  title        = {Rethinking benchmark and contamination for language models with rephrased samples},
  year         = {2023},
  howpublished = {arXiv preprint arXiv:2311.04850},
  note         = {\url{https://arxiv.org/abs/2311.04850}}
}

@inproceedings{kwon-etal-2023,
  author    = {Kwon, W. and Li, Z. and Zhuang, S. and Sheng, Y. and Zheng, L. and Yu, C. H. and Gonzalez, J. E. and Zhang, H. and Stoica, I.},
  title     = {Efficient memory management for large language model serving with {PagedAttention}},
  booktitle = {{Proceedings of the 29th Symposium on Operating Systems Principles (SOSP '23)}},
  publisher = {Association for Computing Machinery},
  year      = {2023},
  note      = {\url{https://doi.org/10.1145/3600006.3613165}}
}

@misc{yanitski-2026-crt-data,
  author       = {Yanitski, D.},
  title        = {Large-language models validate the {C}ognitive {R}eflection {T}est: {D}ata and code},
  year         = {2026},
  type         = {Data set},
  howpublished = {Zenodo},
  note         = {\url{https://doi.org/10.5281/zenodo.24667101}}
}

@misc{yanitski-2026,
  author       = {Yanitski, D.},
  title        = {Reasoning in large language models causes more utilitarian judgments},
  year         = {2026},
  type         = {Manuscript submitted for publication},
  howpublished = {Department of Psychology, University of Alberta}
}

@misc{vllm-2026,
  author       = {{vLLM Project}},
  title        = {{vLLM}: A high-throughput and memory-efficient inference and serving engine for {LLMs}},
  year         = {2026},
  howpublished = {[Computer software]. GitHub},
  note         = {\url{https://github.com/vllm-project/vllm}}
}

@misc{llamacpp-2026,
  author       = {{ggml-org}},
  title        = {llama.cpp: {LLM} inference in {C/C++} ({S}eptember 2026 build)},
  year         = {2026},
  howpublished = {[Computer software]. GitHub},
  note         = {\url{https://github.com/ggml-org/llama.cpp}}
}

@article{fedus-etal-2022,
  author  = {Fedus, W. and Zoph, B. and Shazeer, N.},
  title   = {Switch transformers: {S}caling to trillion parameter models with simple and efficient sparsity},
  journal = {Journal of Machine Learning Research},
  year    = {2022},
  volume  = {23},
  number  = {120},
  pages   = {1--39},
  note    = {\url{https://jmlr.org/papers/v23/21-0998.html}}
}

@misc{qwen-team-2025,
  author       = {{Qwen Team}},
  title        = {{Qwen3} technical report},
  year         = {2025},
  howpublished = {arXiv preprint arXiv:2505.09388},
  note         = {\url{https://arxiv.org/abs/2505.09388}}
}

@incollection{gholami-etal-2022,
  author    = {Gholami, A. and Kim, S. and Dong, Z. and Yao, Z. and Mahoney, M. W. and Keutzer, K.},
  title     = {A survey of quantization methods for efficient neural network inference},
  booktitle = {Low-power computer vision: {I}mprove the efficiency of artificial intelligence},
  editor    = {Thiruvathukal, G. K. and Lu, Y.-H. and Kim, J. and Chen, Y. and Chen, B.},
  publisher = {Chapman and Hall/CRC},
  year      = {2022},
  pages     = {291--326},
  note      = {\url{https://doi.org/10.1201/9781003162810-13}}
}

\newpage
\section{Supplementary Materials}

\setcounter{figure}{0}
\setcounter{table}{0}
\renewcommand{\thefigure}{S\arabic{figure}}
\renewcommand{\thetable}{S\arabic{table}}

\subsection{Full Model Roster}

Tables~\ref{tab:s-roster-1}, \ref{tab:s-roster-2}, and \ref{tab:s-roster-3} list all
\valNComplete{} models in the sample, in alphabetical order of repository identifier. Every
model was served from its published weights on the clusters described in the Method, and no
model was accessed through a hosted service. Parameter counts are total parameters, with
active-per-token parameters for mixture-of-experts models in parentheses (``act.''). The
Switch column gives the chat-template argument that set the reasoning switch. A bare argument
name was sent as \texttt{true} in the reason condition and \texttt{false} in the intuition
condition. Any other argument is shown with its reason-condition value, a slash, and its
intuition-condition value, and a dash marks an argument sent in one condition only. The
\valNReused{} models whose CRT trials come from the earlier collection are marked with a
superscript \emph{r}. The \valNAware{} models flagged by the awareness scan and excluded from
every analysis are marked with a superscript \emph{a}. The remaining \valNAnalyzed{} models
constitute the analysis set.

\begin{table}[p]
\caption{Full Model Roster (Part 1 of 3) of the \valNComplete{} Models in the Sample}
\label{tab:s-roster-1}
\tighttable\scriptsize
\setlength{\tabcolsep}{4pt}
\begin{tabular}{@{}>{\raggedright\arraybackslash}p{3.0in} >{\raggedright\arraybackslash}p{0.85in} >{\raggedright\arraybackslash}p{0.95in} >{\raggedright\arraybackslash}p{1.3in}@{}}
\toprule
Model & Organization & Parameters & Switch \\
\midrule
\texttt{abacusai/\allowbreak Smaug-Mini} & Abacus.AI & 27B & enable\_thinking, reasoning\_effort medium/-- \\
\texttt{ai9stars/\allowbreak G9v3-3B} & AI9Stars & 2.99B & enable\_thinking \\
\texttt{AIDC-AI/\allowbreak Ovis2.5-2B} & AIDC-AI & 3B & enable\_thinking \\
\texttt{AIDC-AI/\allowbreak Ovis2.5-9B} & AIDC-AI & 9B & enable\_thinking \\
\texttt{AIDC-AI/\allowbreak Ovis2.6-30B-A3B} & AIDC-AI & 30B (3B act.) & enable\_thinking \\
\texttt{AIDC-AI/\allowbreak Ovis2.6-80B-A3B}$^{r}$ & AIDC-AI & 80B (3B act.) & enable\_thinking \\
\texttt{aisingapore/\allowbreak Qwen-SEA-LION-v4-32B-IT} & AI Singapore & 32.8B & enable\_thinking \\
\texttt{aisingapore/\allowbreak Qwen-SEA-LION-v4.5-27B-IT}$^{a}$ & AI Singapore & 27.8B & enable\_thinking \\
\texttt{arcee-ai/\allowbreak Homunculus} & Arcee AI & 12.5B & enable\_thinking \\
\texttt{baichuan-inc/\allowbreak Baichuan-M2-32B}$^{r}$ & Baichuan & 33B & thinking\_mode on/off \\
\texttt{CohereLabs/\allowbreak North-Mini-Code-1.0} & Cohere & 30B (3B act.) & reasoning \\
\texttt{deepreinforce-ai/\allowbreak Ornith-1.0-35B}$^{a}$ & deepreinforce-ai & 35B (3B act.) & enable\_thinking \\
\texttt{deepreinforce-ai/\allowbreak Ornith-1.0-397B}$^{a}$ & deepreinforce-ai & 397B (17B act.) & enable\_thinking \\
\texttt{deepreinforce-ai/\allowbreak Ornith-1.0-9B} & deepreinforce-ai & 9B & enable\_thinking \\
\texttt{deepseek-ai/\allowbreak DeepSeek-V3.1} & DeepSeek & 671B (37B act.) & thinking \\
\texttt{deepseek-ai/\allowbreak DeepSeek-V3.1-Terminus} & DeepSeek & 671B (37B act.) & thinking \\
\texttt{deepseek-ai/\allowbreak DeepSeek-V3.2} & DeepSeek & 685B (37B act.) & thinking \\
\texttt{deepseek-ai/\allowbreak DeepSeek-V3.2-Exp} & DeepSeek & 685B (37B act.) & thinking \\
\texttt{deepseek-ai/\allowbreak DeepSeek-V4-Flash}$^{r}$ & DeepSeek & 284B (13B act.) & thinking, reasoning\_effort high/-- \\
\texttt{deepseek-ai/\allowbreak DeepSeek-V4-Flash-0731} & DeepSeek & 284B (13B act.) & thinking, reasoning\_effort low/-- \\
\texttt{deepseek-ai/\allowbreak DeepSeek-V4.1-Flash} & DeepSeek & 552B (16B act.) & thinking, reasoning\_effort 75/-- \\
\texttt{dnotitia/\allowbreak DNA-2.0-14B} & Dnotitia & 14.8B & enable\_thinking \\
\texttt{dnotitia/\allowbreak DNA3.0-0.8B} & Dnotitia & 0.8B & enable\_thinking \\
\texttt{dnotitia/\allowbreak DNA3.0-122B-A10B} & Dnotitia & 122B (10B act.) & enable\_thinking \\
\texttt{dnotitia/\allowbreak DNA3.0-27B}$^{a}$ & Dnotitia & 27.4B & enable\_thinking \\
\texttt{dnotitia/\allowbreak DNA3.0-2B} & Dnotitia & 2B & enable\_thinking \\
\texttt{dnotitia/\allowbreak DNA3.0-35B-A3B} & Dnotitia & 35B (3B act.) & enable\_thinking \\
\texttt{dnotitia/\allowbreak DNA3.0-397B-A17B} & Dnotitia & 397B (17B act.) & enable\_thinking \\
\texttt{dnotitia/\allowbreak DNA3.0-4B} & Dnotitia & 4B & enable\_thinking \\
\texttt{dnotitia/\allowbreak DNA3.0-9B} & Dnotitia & 9B & enable\_thinking \\
\texttt{dots-studio/\allowbreak dots3-note-prev} & Xiaohongshu & 280B (16B act.) & enable\_thinking \\
\texttt{gnani/\allowbreak gnani-evon-v3.3-30B-A3B} & Gnani.ai & 30B (3.5B act.) & enable\_thinking \\
\texttt{google/\allowbreak gemma-4-12B-it} & Google & 11.95B & enable\_thinking \\
\texttt{google/\allowbreak gemma-4-26B-A4B-it} & Google & 25.2B (3.8B act.) & enable\_thinking \\
\texttt{google/\allowbreak gemma-4-31B-it} & Google & 30.7B & enable\_thinking \\
\texttt{google/\allowbreak gemma-4-E2B-it} & Google & 2.3B & enable\_thinking \\
\texttt{google/\allowbreak gemma-4-E4B-it} & Google & 4.5B & enable\_thinking \\
\texttt{iapp/\allowbreak chinda-qwen3-4b} & iApp Technology & 4.02B & enable\_thinking \\
\texttt{ibm-granite/\allowbreak granite-4.2-30b} & IBM & 30B & enable\_thinking \\
\texttt{ibm-granite/\allowbreak granite-4.2-3b} & IBM & 3B & enable\_thinking \\
\texttt{ibm-granite/\allowbreak granite-4.2-8b}$^{a}$ & IBM & 8B & enable\_thinking \\
\texttt{internlm/\allowbreak Intern-S2-397B} & InternLM & 397B (17B act.) & enable\_thinking \\
\texttt{internlm/\allowbreak Intern-S2-Mobius} & InternLM & 35B (3B act.) & enable\_thinking \\
\texttt{internlm/\allowbreak Intern-S2-Preview}$^{a}$ & InternLM & 35B & enable\_thinking \\
\texttt{KORMo-Team/\allowbreak KORMo-10B-sft} & KAIST & 10.8B & enable\_thinking \\
\texttt{Kwaipilot/\allowbreak KAT-Coder-V2.5-Dev} & Kwaipilot & 35B (3B act.) & enable\_thinking \\
\bottomrule

\end{tabular}
\caption*{\textit{Note}. act.\ = active parameters per token. $^{r}$CRT trials from the
earlier collection. $^{a}$Excluded by the awareness scan.}
\end{table}

\begin{table}[p]
\caption{Full Model Roster (Part 2 of 3) of the \valNComplete{} Models in the Sample}
\label{tab:s-roster-2}
\tighttable\scriptsize
\setlength{\tabcolsep}{4pt}
\begin{tabular}{@{}>{\raggedright\arraybackslash}p{3.0in} >{\raggedright\arraybackslash}p{0.85in} >{\raggedright\arraybackslash}p{0.95in} >{\raggedright\arraybackslash}p{1.3in}@{}}
\toprule
Model & Organization & Parameters & Switch \\
\midrule
\texttt{LGAI-EXAONE/\allowbreak EXAONE-4.0-1.2B}$^{a}$ & LG AI Research & 1.2B & enable\_thinking \\
\texttt{LGAI-EXAONE/\allowbreak EXAONE-4.0-32B}$^{r}$ & LG AI Research & 32B & enable\_thinking \\
\texttt{LGAI-EXAONE/\allowbreak EXAONE-4.5-33B} & LG AI Research & 33B & enable\_thinking \\
\texttt{LGAI-EXAONE/\allowbreak K-EXAONE-236B-A23B} & LG AI Research & 236B (23B act.) & enable\_thinking \\
\texttt{mistralai/\allowbreak Mistral-Medium-3.5-128B} & Mistral AI & 128B & reasoning\_effort high/none \\
\texttt{moonshotai/\allowbreak Kimi-K2.5}$^{a}$ & Moonshot AI & 1T (32B act.) & thinking \\
\texttt{moonshotai/\allowbreak Kimi-K2.6}$^{r}$ & Moonshot AI & 1T (32B act.) & thinking \\
\texttt{Nanbeige/\allowbreak Nanbeige4.2-3B}$^{a}$ & BOSS Zhipin (Nanbeige Lab) & 4B & enable\_thinking \\
\texttt{naver-hyperclovax/\allowbreak HyperCLOVAX-SEED-Think-14B}$^{r}$ & Naver & 14.74B & force\_reasoning true/--, skip\_reasoning --/true \\
\texttt{naver-hyperclovax/\allowbreak HyperCLOVAX-SEED-Think-32B}$^{r}$ & Naver & 32B & skip\_reasoning false/true \\
\texttt{nex-agi/\allowbreak Nex-N2-mini} & nex-agi & 35B (3B act.) & enable\_thinking \\
\texttt{nex-agi/\allowbreak Nex-N2-Pro} & nex-agi & 397B (17B act.) & enable\_thinking \\
\texttt{nex-agi/\allowbreak Nex-N2.5-mini} & nex-agi & 35B (3B act.) & reasoning\_effort high/none \\
\texttt{nex-agi/\allowbreak Nex-N2.5-Pro} & nex-agi & 397B (17B act.) & reasoning\_effort high/none \\
\texttt{nvidia/\allowbreak Nemotron-3-Nano-Omni-30B-A3B-Reasoning-BF16}$^{r}$ & NVIDIA & 31B (3B act.) & enable\_thinking \\
\texttt{nvidia/\allowbreak Nemotron-Cascade-2-30B-A3B}$^{r}$ & NVIDIA & 30B (3B act.) & enable\_thinking \\
\texttt{nvidia/\allowbreak NVIDIA-Nemotron-3-Nano-30B-A3B-BF16} & NVIDIA & 30B (3.5B act.) & enable\_thinking \\
\texttt{nvidia/\allowbreak NVIDIA-Nemotron-3-Nano-4B-BF16} & NVIDIA & 4B & enable\_thinking \\
\texttt{nvidia/\allowbreak NVIDIA-Nemotron-3-Super-120B-A12B-BF16}$^{r}$ & NVIDIA & 120B (12B act.) & enable\_thinking \\
\texttt{nvidia/\allowbreak NVIDIA-Nemotron-3-Ultra-550B-A55B-BF16} & NVIDIA & 550B (55B act.) & enable\_thinking \\
\texttt{nvidia/\allowbreak NVIDIA-Nemotron-3.5-Lightning-30B-A3B-BF16}$^{a}$ & NVIDIA & 30B (3B act.) & enable\_thinking \\
\texttt{nvidia/\allowbreak NVIDIA-Nemotron-Labs-3-Elastic-30B-A3B-BF16} & NVIDIA & 30B (3.6B act.) & enable\_thinking \\
\texttt{nvidia/\allowbreak NVIDIA-Nemotron-Labs-3-Puzzle-75B-A9B-BF16} & NVIDIA & 75.3B (9.3B act.) & enable\_thinking \\
\texttt{openbmb/\allowbreak MiniCPM-V-4.6} & OpenBMB & 1B & enable\_thinking \\
\texttt{openbmb/\allowbreak MiniCPM-V-4\_5} & OpenBMB & 8B & enable\_thinking \\
\texttt{openbmb/\allowbreak MiniCPM4.1-8B} & OpenBMB & 8B & enable\_thinking \\
\texttt{openbmb/\allowbreak MiniCPM5-1B} & OpenBMB & 1B & enable\_thinking \\
\texttt{openbmb/\allowbreak MiniCPM5-2B} & OpenBMB & 2.52B & enable\_thinking \\
\texttt{ornith-ai/\allowbreak Ornith-1.5-35B-A3B} & deepreinforce-ai & 35B (3B act.) & enable\_thinking \\
\texttt{ornith-ai/\allowbreak Ornith-1.5-397B} & deepreinforce-ai & 397B (17B act.) & enable\_thinking \\
\texttt{ornith-ai/\allowbreak Ornith-1.5-9B} & deepreinforce-ai & 9B & enable\_thinking \\
\texttt{poolside/\allowbreak Laguna-M.1} & poolside & 225B (23B act.) & enable\_thinking \\
\texttt{poolside/\allowbreak Laguna-XS-2.1} & poolside & 33B (3B act.) & enable\_thinking \\
\texttt{poolside/\allowbreak Laguna-XS.2} & poolside & 33B (3B act.) & enable\_thinking \\
\texttt{Qwen/\allowbreak Qwen3-0.6B} & Alibaba (Qwen) & 0.6B & enable\_thinking \\
\texttt{Qwen/\allowbreak Qwen3-1.7B} & Alibaba (Qwen) & 1.7B & enable\_thinking \\
\texttt{Qwen/\allowbreak Qwen3-14B} & Alibaba (Qwen) & 14.8B & enable\_thinking \\
\texttt{Qwen/\allowbreak Qwen3-235B-A22B}$^{r}$ & Alibaba (Qwen) & 235B (22B act.) & enable\_thinking \\
\texttt{Qwen/\allowbreak Qwen3-30B-A3B}$^{r}$ & Alibaba (Qwen) & 30.5B (3.3B act.) & enable\_thinking \\
\texttt{Qwen/\allowbreak Qwen3-32B}$^{a}$ & Alibaba (Qwen) & 32.8B & enable\_thinking \\
\texttt{Qwen/\allowbreak Qwen3-4B} & Alibaba (Qwen) & 4B & enable\_thinking \\
\texttt{Qwen/\allowbreak Qwen3-8B} & Alibaba (Qwen) & 8.2B & enable\_thinking \\
\texttt{Qwen/\allowbreak Qwen3.5-0.8B}$^{a}$ & Alibaba (Qwen) & 0.8B & enable\_thinking \\
\texttt{Qwen/\allowbreak Qwen3.5-122B-A10B}$^{r}$ & Alibaba (Qwen) & 122B (10B act.) & enable\_thinking \\
\texttt{Qwen/\allowbreak Qwen3.5-27B}$^{a}$ & Alibaba (Qwen) & 27B & enable\_thinking \\
\texttt{Qwen/\allowbreak Qwen3.5-2B}$^{a}$ & Alibaba (Qwen) & 2B & enable\_thinking \\
\bottomrule

\end{tabular}
\caption*{\textit{Note}. act.\ = active parameters per token. $^{r}$CRT trials from the
earlier collection. $^{a}$Excluded by the awareness scan.}
\end{table}

\begin{table}[p]
\caption{Full Model Roster (Part 3 of 3) of the \valNComplete{} Models in the Sample}
\label{tab:s-roster-3}
\tighttable\scriptsize
\setlength{\tabcolsep}{4pt}
\begin{tabular}{@{}>{\raggedright\arraybackslash}p{3.0in} >{\raggedright\arraybackslash}p{0.85in} >{\raggedright\arraybackslash}p{0.95in} >{\raggedright\arraybackslash}p{1.3in}@{}}
\toprule
Model & Organization & Parameters & Switch \\
\midrule
\texttt{Qwen/\allowbreak Qwen3.5-35B-A3B} & Alibaba (Qwen) & 35B (3B act.) & enable\_thinking \\
\texttt{Qwen/\allowbreak Qwen3.5-397B-A17B}$^{a}$ & Alibaba (Qwen) & 397B (17B act.) & enable\_thinking \\
\texttt{Qwen/\allowbreak Qwen3.5-4B} & Alibaba (Qwen) & 4B & enable\_thinking \\
\texttt{Qwen/\allowbreak Qwen3.5-9B}$^{r}$ & Alibaba (Qwen) & 9B & enable\_thinking \\
\texttt{Qwen/\allowbreak Qwen3.6-27B}$^{a}$ & Alibaba (Qwen) & 27B & enable\_thinking \\
\texttt{Qwen/\allowbreak Qwen3.6-35B-A3B}$^{a}$ & Alibaba (Qwen) & 35B (3B act.) & enable\_thinking \\
\texttt{Qwen/\allowbreak Qwen3.8-27B} & Alibaba (Qwen) & 27B & enable\_thinking, reasoning\_effort medium/-- \\
\texttt{Qwen/\allowbreak Qwen3.8-Flash-Next}$^{a}$ & Alibaba (Qwen) & 125B (6B act.) & enable\_thinking, reasoning\_effort medium/-- \\
\texttt{RefalMachine/\allowbreak RuadaptQwen3-32B-Instruct} & Ruadapt & 32.8B & enable\_thinking \\
\texttt{RefalMachine/\allowbreak RuadaptQwen3-4B-Hybrid} & Ruadapt & 4.02B & enable\_thinking \\
\texttt{RefalMachine/\allowbreak RuadaptQwen3-8B-Hybrid} & Ruadapt & 8.14B & enable\_thinking \\
\texttt{saai-sa/\allowbreak ASL-4B-v1} & SAAI & 4.66B & enable\_thinking \\
\texttt{shuttleai/\allowbreak shuttle-3.5}$^{a}$ & ShuttleAI & 32.8B & enable\_thinking \\
\texttt{t-tech/\allowbreak T-pro-it-2.0} & T-Technologies & 32.8B & enable\_thinking \\
\texttt{tencent/\allowbreak Hunyuan-0.5B-Instruct} & Tencent & 0.5B & enable\_thinking \\
\texttt{tencent/\allowbreak Hunyuan-1.8B-Instruct} & Tencent & 1.8B & enable\_thinking \\
\texttt{tencent/\allowbreak Hunyuan-4B-Instruct} & Tencent & 4B & enable\_thinking \\
\texttt{tencent/\allowbreak Hy3} & Tencent & 295B (21B act.) & reasoning\_effort high/no\_think \\
\texttt{tencent/\allowbreak Youtu-LLM-2B} & Tencent & 1.96B & enable\_thinking, forced\_thinking true/-- \\
\texttt{thinkingmachines/\allowbreak Inkling-Small} & Thinking Machines Lab & 276B (12B act.) & reasoning\_effort high/none \\
\texttt{thomsonreuters/\allowbreak Thomson-1.0-Small}$^{a}$ & Thomson Reuters & 35B (3B act.) & enable\_thinking \\
\texttt{TokenRhythm/\allowbreak NeoHorse-1-4B} & TokenRhythm & 4.21B & enable\_thinking \\
\texttt{TokenRhythm/\allowbreak NeoHorse-1-9B} & TokenRhythm & 8.95B & enable\_thinking \\
\texttt{Trendyol/\allowbreak Trendyol-LLM-8B-T1} & Trendyol & 8.2B & enable\_thinking \\
\texttt{XHToken/\allowbreak Spark-X2.5-1.7B}$^{a}$ & XHToken & 1.7B & enable\_thinking \\
\texttt{XHToken/\allowbreak Spark-X2.5-4B} & XHToken & 4.11B & enable\_thinking \\
\texttt{XiaomiMiMo/\allowbreak MiMo-V2-Flash}$^{a}$ & Xiaomi & 309B (15B act.) & enable\_thinking \\
\texttt{XiaomiMiMo/\allowbreak MiMo-V2.5}$^{r}$ & Xiaomi & 310B (15B act.) & enable\_thinking \\
\texttt{XiaomiMiMo/\allowbreak MiMo-V2.5-Pro}$^{r}$ & Xiaomi & 1.02T (42B act.) & enable\_thinking \\
\texttt{XiaomiMiMo/\allowbreak MiMo-V2.6-Flash-RL} & Xiaomi & 309B (15B act.) & enable\_thinking \\
\texttt{XiaomiMiMo/\allowbreak MiMo-V2.6-Pro-RL} & Xiaomi & 1.02T (42B act.) & enable\_thinking \\
\texttt{YannQi/\allowbreak R-4B} & YannQi & 5B & thinking\_mode long/short \\
\texttt{zai-org/\allowbreak GLM-4.5} & Z.AI & 355B (32B act.) & enable\_thinking \\
\texttt{zai-org/\allowbreak GLM-4.5-Air}$^{r}$ & Z.AI & 106B (12B act.) & enable\_thinking \\
\texttt{zai-org/\allowbreak GLM-4.5V}$^{r}$ & Z.AI & 106B (12B act.) & enable\_thinking \\
\texttt{zai-org/\allowbreak GLM-4.6} & Z.AI & 357B (32B act.) & enable\_thinking \\
\texttt{zai-org/\allowbreak GLM-4.6V}$^{r}$ & Z.AI & 106B (12B act.) & enable\_thinking \\
\texttt{zai-org/\allowbreak GLM-4.6V-Flash} & Z.AI & 10.3B & enable\_thinking \\
\texttt{zai-org/\allowbreak GLM-4.7}$^{a}$ & Z.AI & 358B (32B act.) & enable\_thinking \\
\texttt{zai-org/\allowbreak GLM-4.7-Flash} & Z.AI & 31B (3B act.) & enable\_thinking \\
\texttt{zai-org/\allowbreak GLM-5}$^{r}$ & Z.AI & 744B (40B act.) & enable\_thinking \\
\texttt{zai-org/\allowbreak GLM-5.1}$^{r}$ & Z.AI & 754B (40B act.) & enable\_thinking \\
\texttt{zai-org/\allowbreak GLM-5.2} & Z.AI & 753B & enable\_thinking \\
\texttt{Zhihu-ai/\allowbreak Zhi-Create-Qwen3-32B} & Zhihu & 32.8B & enable\_thinking \\
\bottomrule

\end{tabular}
\caption*{\textit{Note}. act.\ = active parameters per token. $^{r}$CRT trials from the
earlier collection. $^{a}$Excluded by the awareness scan.}
\end{table}

\subsection{Verbatim Prompt}

\textit{Prompt construction.} Every CRT trial sent a single user message consisting of
the item's vignette, its question, and the response instruction, joined by blank
lines, in that order:

\begin{verbatim}
{vignette}

{question}

Respond with ONLY a JSON object of the form {"answer": N} where N is
the numeric answer to the problem (a number only, with no units or
other text).
\end{verbatim}

\noindent
The response instruction was identical for all eight items and both conditions, and the
user message did not differ between conditions. No system message was sent in either
condition. The reasoning toggle was set through the chat-template argument listed for each
model in Tables~\ref{tab:s-roster-1}--\ref{tab:s-roster-3}, so the condition changed only
the control markup that the model's own template wrote around this message (and, for
\valNsettingsAn{} analyzed models, a single reasoning-settings line, as described in the
Method). The eight vignette--question pairs are reproduced verbatim in
Table~\ref{tab:crt-items} of the main text.

\subsection{Item Identifiers and Published Response Rates}

Table~\ref{tab:s-items} maps the internal item identifiers used in the data files
(\texttt{crt\_01}--\texttt{crt\_08}) to the short item labels used in the Results
figures and tables, together with each item's correct answer, its intuitive lure
(the published most common error [MCE]), and the published human first-exposure
solution and MCE rates from Table~1 of \citet{meyer-etal-2024}. Each trial presented a single
item. The verbatim vignette and question text, as administered, appear in
Table~\ref{tab:crt-items} of the main text. Items 1, 3, and 6 are the
updated variants of the three original CRT items (bat-and-ball, lily pads, and
machines-and-widgets, respectively) \citep{frederick-2005, meyer-etal-2024}. Items 5
and 7 recast the class-rank and barrel items of an earlier expanded CRT in new surface
content, with the same correct answers and lures \citep{toplak-etal-2014}. Items 2, 4, and
8 complete the set of \citet{meyer-etal-2024}.

\begin{table}[tbp]
\caption{CRT Item Identifiers, Labels, Answer Keys, and Published Human Response Rates}
\label{tab:s-items}
\centering
\begin{tabular}{@{}l l r r r r@{}}
\toprule
Item ID & Label & Correct & Lure (MCE) & Sol.\ rate & MCE rate \\
\midrule
\texttt{crt\_01} & Drill \& hammer    & \valItemOneCorrect   & \valItemOneLure   & \valItemOneSol   & \valItemOneMCE   \\
\texttt{crt\_02} & Dog \& cat         & \valItemTwoCorrect   & \valItemTwoLure   & \valItemTwoSol   & \valItemTwoMCE   \\
\texttt{crt\_03} & Doubling bird      & \valItemThreeCorrect & \valItemThreeLure & \valItemThreeSol & \valItemThreeMCE \\
\texttt{crt\_04} & Toaster 20\% off   & \valItemFourCorrect  & \valItemFourLure  & \valItemFourSol  & \valItemFourMCE  \\
\texttt{crt\_05} & 15th tallest       & \valItemFiveCorrect  & \valItemFiveLure  & \valItemFiveSol  & \valItemFiveMCE  \\
\texttt{crt\_06} & Elves \& gifts     & \valItemSixCorrect   & \valItemSixLure   & \valItemSixSol   & \valItemSixMCE   \\
\texttt{crt\_07} & Cider (Jack/Jill)  & \valItemSevenCorrect & \valItemSevenLure & \valItemSevenSol & \valItemSevenMCE \\
\texttt{crt\_08} & Green apples       & \valItemEightCorrect & \valItemEightLure & \valItemEightSol & \valItemEightMCE \\
\bottomrule
\end{tabular}
\caption*{\textit{Note}. Correct = the item's single correct numeric answer. Lure
(MCE) = the published most common error. Sol.\ rate and MCE rate are the published
human first-exposure frequencies of the correct answer and the MCE
\citep{meyer-etal-2024}. Scoring matched a trial's parsed numeric answer, after rounding to the nearest integer,
exactly against the Correct and Lure values.}
\end{table}

\subsection{Awareness-Signal Definitions}

The awareness scan read the full stored reply, reasoning field included, of every
attempt of every CRT trial, invalid attempts included, and matched case-insensitive
patterns over the lower-cased text. The scan covered every attempt of the \valScanTrials{} CRT
trials from the \valScanModels{} models in the sample (\valScanCharsM{} million characters of
text). A model was flagged as aware if any counted signal appeared on any of
its trials.
Table~\ref{tab:s-aware-signals} defines the five counted signals (A--E) and the three
structural signals (X1--X3) that were detected and reported for transparency but
deliberately \emph{not} counted, with the number of models triggering each.

\begin{table}[tbp]
\caption{Counted Awareness Signals (A--E) and Excluded Structural Signals (X1--X3)}
\label{tab:s-aware-signals}
\centering
{\renewcommand{\baselinestretch}{1}\small
\begin{tabular}{@{}c p{2.2in} p{2.6in} r@{}}
\toprule
Key & Signal & Definition & Models \\
\midrule
A & Names the test & Mentions ``cognitive reflection'' or ``CRT'' & \valSigA/\valScanModels \\
B & Original-CRT reference & Connects an item to one of the three original CRT items \citep{frederick-2005} (bat-and-ball, machines and widgets co-occurring, or lily pads / doubling pond) & \valSigB/\valScanModels \\
C & Authors / this study & Names the authors, the 2024 study or its title or venue, or the CRT's foundational literature & \valSigC/\valScanModels \\
D & Documented MCE & Attributes the item's \emph{published} MCE value \citep{meyer-etal-2024} to people (population attribution plus response or error verb, and value matched with digit boundaries) & \valSigD/\valScanModels \\
E & Prior encounter & Explicit claim of having encountered the item or its source (e.g., ``standard CRT question'') & \valSigE/\valScanModels \\
\midrule
X1 & Trick suspicion & ``Is this a trick?'', ``designed to mislead'' (the reaction the lure is engineered to provoke) & \valStructTrick/\valScanModels \\
X2 & Generic template & ``Classic/famous riddle or puzzle'' with no CRT link (recognizes a genre, not the test) & \valStructTemplate/\valScanModels \\
X3 & Lure prediction & ``The intuitive/tempting answer would be $X$'' with no claim about what people actually do & \valStructLure/\valScanModels \\
\bottomrule
\end{tabular}
}
\caption*{\textit{Note}. A model can trigger several signals. Signals X1--X3 never
contribute to the awareness count. The \valStructOnly{} models that showed \emph{only} structural signals
were retained. Signal D separates \emph{reporting} the documented human error
(counted) from \emph{predicting} a plausible naive answer (structural, X3): the
matched value must equal the item's published MCE, with digit boundaries so that,
e.g., the MCE ``30'' cannot be triggered by the ``330'' or ``300'' in the item's own
arithmetic, and the value must be attributed to people (``most people mistakenly
answer 30''), not merely flagged as tempting (``the intuitive answer is 30'').}
\end{table}

The dividing line follows the source paper's own account of the items: each is
engineered around ``a dominant intuitive lure which tempts respondents into performing
some simple, but inappropriate, operation'' \citep{meyer-etal-2024}, so a careful
solver \emph{should} pause, suspect a trick, and notice the tempting answer.
These reactions are derivable from the item's structure and therefore excluded. What
cannot be derived from structure is knowledge of the item as a studied stimulus: the
test's name, its lineage, its authors, what people empirically answer, or a prior
encounter. Table~\ref{tab:s-aware-models} lists the \valScanAware{} flagged models and their
triggered signals. Notably, signal C (the cleanest test of whether the 2024 study
itself is in training data) fired for no model, and signal B, where recognition
did occur, pointed at the classic CRT lineage rather than the present items. The scan
is a conservative lower bound (it sees only verbalized awareness, and the intuition
condition normally emits little or no reasoning text), and it is a pattern-matching
heuristic, so the full report archives verbatim matched snippets for every flagged signal
of every flagged model, and every
exclusion can be audited by hand against the archived data.

\begin{table}[tbp]
\caption{The \valScanAware{} Awareness-Flagged Models and Their Triggered Signals}
\label{tab:s-aware-models}
\centering
{\renewcommand{\baselinestretch}{1}\small
\begin{tabular}{@{}l l@{}}
\toprule
Model & Signals \\
\midrule
\texttt{aisingapore/Qwen-SEA-LION-v4.5-27B-IT} & D \\
\texttt{deepreinforce-ai/Ornith-1.0-35B} & A, D \\
\texttt{deepreinforce-ai/Ornith-1.0-397B} & D \\
\texttt{dnotitia/DNA3.0-27B} & B \\
\texttt{ibm-granite/granite-4.2-8b} & D \\
\texttt{internlm/Intern-S2-Preview} & D \\
\texttt{LGAI-EXAONE/EXAONE-4.0-1.2B} & D \\
\texttt{moonshotai/Kimi-K2.5} & D \\
\texttt{Nanbeige/Nanbeige4.2-3B} & B \\
\texttt{nvidia/NVIDIA-Nemotron-3.5-Lightning-30B-A3B-BF16} & D \\
\texttt{Qwen/Qwen3-32B} & D \\
\texttt{Qwen/Qwen3.5-0.8B} & E \\
\texttt{Qwen/Qwen3.5-27B} & B, D \\
\texttt{Qwen/Qwen3.5-2B} & D \\
\texttt{Qwen/Qwen3.5-397B-A17B} & A, B, D \\
\texttt{Qwen/Qwen3.6-27B} & A, B \\
\texttt{Qwen/Qwen3.6-35B-A3B} & A, B, D \\
\texttt{Qwen/Qwen3.8-Flash-Next} & D \\
\texttt{shuttleai/shuttle-3.5} & D \\
\texttt{thomsonreuters/Thomson-1.0-Small} & A, B, D \\
\texttt{XHToken/Spark-X2.5-1.7B} & D, E \\
\texttt{XiaomiMiMo/MiMo-V2-Flash} & D \\
\texttt{zai-org/GLM-4.7} & B \\
\bottomrule

\end{tabular}
}
\caption*{\textit{Note}. Signal keys as in Table~\ref{tab:s-aware-signals}. All \valScanAware{}
models were excluded from every analysis. Verbatim evidence snippets for each flag are
archived with the study data.}
\end{table}

\subsection{Truncation, Validity, and Ceiling Accounting}

\textit{Row selection.} Every attempt of every trial was appended as one row to an
append-only JSON-lines file per model, and rows were never edited or overwritten. Dataset
assembly kept one row per (model, item, condition, repetition) trial. A valid row was
preferred over an invalid one, and among valid rows the one with the lowest attempt
number was kept. Newly collected trials were retried up to three attempts in all, and
trials from the earlier collection up to two.

\textit{Truncation.} A reply cut off at the output limit is not interpretable as an
answer. The new runs recorded the reason each reply ended, and a reply that ended at the
limit was marked invalid and retried. The earlier collection did not record that reason,
so a model's operative limit was detected from its own output distribution. When the
model's maximum completion-token count among valid replies was at least 4,000 tokens and
occurred on at least three trials, which is the signature of replies pinned at a hard
limit rather than ending naturally, all rows at that maximum were treated as cut off. No
model from the earlier collection showed this signature.

\textit{Completeness accounting.} All \valNComplete{} models in the sample had a row for
every one of the \valNSampleTrials{} CRT trials (\valNTrialsPerModel{} per model). The
\valNTruncSample{} trials that remained cut off after the last attempt all came from
\valNTruncModels{} model, which the awareness scan excluded. The \valNAnalyzed{} analyzed
models therefore contribute $\valNAnalyzed \times \valNTrialsPerModel = \valNTrials$ valid
trials, and every paired comparison rests on the same \valNTrialsPerModel-trial design.

\textit{Ceiling accounting.} As a robustness note on the scale analysis
(Figure~\ref{fig:crt-scale}), \valNCeilingBoth{} of the \valNAnalyzed{} analyzed models,
concentrated among the largest, were at perfect accuracy in both conditions, and all
\valNCeilingBoth{} also produced zero lure responses in both conditions. Their measured
reasoning effect is structurally zero. No model that was perfect under intuition declined
under reason. All \valLureTie{} models whose lure rate did not change between conditions
produced no lure response in either condition. The across-model effects reported in the
Results therefore carry these models as zeros, understating rather than inflating the
effect of deliberation among models with headroom. The headroom-restricted and normalized
effects, the recomputed scale correlations, and the organization-collapsed direction
tests appear in Robustness to Ceiling and Floor Effects in the main text, and the full
computation is archived with the study data.

\end{document}